\documentclass[10pt,journal]{IEEEtran}
\usepackage{mathrsfs}
\usepackage{amsmath}
\usepackage{url}

\ifCLASSOPTIONcompsoc
\else
\fi
\ifCLASSINFOpdf
\else
\fi
\usepackage{graphicx}
\usepackage{subfigure}
\usepackage{cite}
\newtheorem{theorem}{Theorem}

\newtheorem{lemma}{Lemma}
\newtheorem{remark}{Remark}
\newtheorem{proposition}{Proposition}

\usepackage{psfrag}
\usepackage{txfonts}
\usepackage{latexsym,bm}
\usepackage{color}
\usepackage{float}
\usepackage[table]{xcolor}
\usepackage{multirow}
\usepackage{graphicx}
\usepackage{subfigure}

\usepackage{color,xcolor}

\begin{document}

\title{Learning through deterministic assignment  of hidden parameters}

\author{Jian Fang,~Shaobo Lin, and~Zongben Xu
\IEEEcompsocitemizethanks{\IEEEcompsocthanksitem  J. Fang and Z. Xu
are with the School of Mathematics and Statistics, Xi'an
        Jiaotong University, Xi'an 710048, China.  S. Lin is with the
Department of Statistics, Wenzhou University, Wenzhou 325035, China.
 }
 \thanks{Shaobo Lin is the corresponding author (sblin1983@gmail.com).}}

\IEEEcompsoctitleabstractindextext{%
\begin{abstract}
Supervised learning frequently  boils down to determining hidden and
bright parameters in a parameterized hypothesis space based on
finite
 input-output samples. The hidden parameters determine the   nonlinear mechanism of an estimator, while the
bright parameters characterize  the linear mechanism. In traditional
learning paradigm,
 hidden and bright parameters are not distinguished and trained
simultaneously in one learning process. Such an one-stage learning
(OSL) brings a benefit of theoretical analysis but   suffers
  from the  high computational burden.
In this paper, we propose a two-stage learning (TSL) scheme,
learning through deterministic assignment of hidden parameters
(LtDaHP), where we suggest to deterministically generate the hidden
parameters by using
    minimal Riesz energy points on a sphere and
  equally spaced points in an interval. We theoretically show that
with such deterministic assignment of hidden parameters, LtDaHP with
a neural network realization  almost shares the same generalization
performance with that of OSL. Then, LtDaHP provides an effective way
to overcome the high computational burden of OSL. We present a
series of simulations and application examples to support the
 outperformance of LtDaHP.
\end{abstract}


}

\maketitle

\IEEEdisplaynotcompsoctitleabstractindextext

\IEEEpeerreviewmaketitle

\section{Introduction}

In physical or biological systems, engineering applications,
financial studies, and many other fields, only can finite number of
samples  be obtained. Supervised learning aims at synthesizing a
function (or mapping)  to represent or approximate an unknown but
definite relation  between the input and output, based on the
input-output samples. The learning process is accomplished with the
selection of   a  hypothesis space and   a learning  algorithm. The
hypothesis space is a family of functions endowed with certain
structures, very often, a space spanned by a set of parameterized
functions like
\begin{equation}\label{general hypothesis space}
           \mathcal{H}=span\left\{ \psi (\xi _{j},x):\xi _{j}\in \Omega
                ,j=1,2,\dots,N\right\}
\end{equation}%
where $\xi _{j}$ is a parameter  for specifying the $j$-th function
$\psi _{j}(\cdot):=\psi(\xi_j,\cdot)$ and $\Omega $ is a class of
parameters. A  typical example  is  the three-layer  feed-forward
neural networks (FNNs) in which $\psi(\xi _{j},x)$ is the response
of the $j$-th neuron in the hidden layer with $\xi_{j}$ being all
the connection weights connected to the neurons \cite{Hagan1996}. A
learning algorithm  is then defined by some optimization scheme to
derive
 an estimator in $\mathcal H$ based on the given samples.
To distinguish the type of parameters, we call each
$\psi_{j}(\cdot):=\psi(\xi _{j},\cdot)$ a \textit{hidden
predictor}, $\xi_{j}$ a\textit{\ hidden parameter}, and $%
 a_{j}$ a \textit{bright parameter}. Then, for a nonlinear function $\psi$, it follows
from (\ref{general hypothesis space}) that  \textit{hidden
parameters} determine the attributions of \textit{hidden predictors} of the estimator (
nonlinear mechanism), while  \textit{bright
parameters} characterize how
  \textit{hidden predictors} are linearly combined (linear
mechanism). In this sense, supervised learning
boils down to determining   \textit{hidden and bright parameters} in
a parameterized hypothesis space.

In traditional learning paradigm,  \textit{hidden and bright
parameters} are not distinguished and trained simultaneously. Such a scheme is featured as the \textit{one-stage
learning} (OSL). The well known support vector machine (SVM)
\cite{Taylor2004}, kernel ridge regression \cite{Cucker2007} and FNNs \cite{Hagan1996}   are   typical examples
of the OSL scheme. OSL has a benefit of theoretical attractiveness
in the sense that this  scheme  enables to realize the optimal
generalization error bounds
\cite{Maiorov2006a,Steinwart2009,Lin2018}.
However, it inevitably requires  to solve some nonlinear optimization
problem, which usually   suffers  from
the time-consuming difficulty, especially for   problems with
large-sized samples.

To circumvent this difficulty, a \textit{two-stage learning} (TSL)
scheme, featured as \textit{learning through random assignment of
hidden parameters} (LtRaHP), was developed and widely used
\cite{Cao2015,Huang2006,Jaeger2004,Lowe1989,Pao1989} in the last two
decades. LtRaHP  assigns randomly the \textit{hidden parameters} in
the first stage and determines the \textit{bright parameters} by
solving a linear least-square  problem in the second stage. Typical
examples of LtRaHP include, the random vector functional-link
networks (RVFLs) \cite{Pao1989}, the echo-state neural networks
(ESNs) \cite{Jaeger2004}, the random weight neural networks (RWNNs)
\cite{Cao2015} and the extreme learning machine (ELM)
\cite{Huang2006}. LtRaHP significantly reduces the computational
burden of OSL without sacrificing the prediction accuracy very much,
as partially justified   in our recent theoretical studies
\cite{Lin2014,Liu2013}. However, due to the randomness of the
\textit{hidden parameters}, a satisfactory  generalization
capability of LtRaHP is achieved only in the sense of expectation.
This then leads to an uncertainty problem: it is uncertain whether a
single trail of the scheme succeeds or not. Consequently, to yield a
convincing result,
 multiple times of trails are required in the  training process of LtRaHP.

From these studies, we   draw a simple conclusion on the pros and
cons of   existing learning schemes.   OSL possesses promising
generalization capabilities but it is built on the high
computational burden, while LtRaHP has charming computational
advantages but it suffers from an uncertainty problem. Thus, it is
still open  to find an efficient and feasible learning scheme,
especially when the size of data is huge. Our aim in the present
paper is to develop a new   TSL scheme. Our core idea is to apply a
deterministic mechanism for the assignment of \textit{hidden
parameters} in place of the random assignment in LtRaHP.
Accordingly, the new TSL scheme will be featured as \textit{learning
through deterministic assignment of hidden parameters} (LtDaHP). We
will show that LtDaHP outperforms LtRaHP in the sense that LtDaHP
avoids the uncertainty problem of LtRaHP without increasing the
computational complexity.

As the popularity of neural networks in recent years \cite{Goodfellow,Peng2018,Chen2018,Zhu2018}, we
equip the LtDaHP scheme with an FNN-instance to show its
outperformance.  Taking  inner weights as   minimal
Riesz energy points on a sphere and     thresholds as
equally spaced points (ESPs) in an interval, we can define an
FNN-realization of LtDaHP. We theoretically justify that so
defined LtDaHP outperforms both LtRaHP and OSL in many ways.
Firstly, LtDaHP can achieve the almost optimal generalization error
bounds   of the OSL schemes;
  Secondly,
LtDaHP significantly  reduces the computational burden of OSL;
Finally, unlike LtRaHP, LtDaHP can find a satisfactory estimator in
a single time of trial.  Thus, LtDaHP provides an effective way of
overcoming both the high computational burden difficulty of OSL and
the uncertainty problem of LtRaHP. We also  provide a series of
simulations and application examples to support the outperformance
of LtDaHP.

The rest of paper is organized  as follows. Section II aims at introducing the new FNN-realization of LtDaHP as well as  a
brief introduction of the minimal Riesz energy configuration problem
on the  sphere. In Section III, we verify the almost
optimality of LtDaHP in the framework of statistical learning
theory. In Section IV, we provide the simulations and application
examples to support the outperformance of LtDaHP and the correctness
of the theoretical assertions we have made.    We conclude the paper
 in Section V with some   remarks.

\section{FNN-Realization of LtDaHP}

In this section, after providing the motivation of the LtDaHP scheme and briefly introducing    minimal Riesz energy points on the sphere,
we formalize an FNN-realization of LtDaHP.

\subsection{Motivations}

FNNs, taking  three-layer FNNs with one output neuron for
example, look for the estimators of the form
$f_{FNN}(x)=\sum_{j=1}^{N}a_{j}\phi (\alpha_{j}\cdot x-b_j)$ where
$\alpha_{j}$ is the inner weight which connects  the input layer
to the $j$-th hidden neuron, $b_{j}$ is the threshold
 of the $j$-th hidden neuron, $\phi $ is the nonlinear activation
function, and $a_{j}$ is the outer weights that connects
 the $j$-th hidden layer to the
output layer. In FNNs, the \textit{hidden parameters} are $\{\alpha_{j},b_{j}%
\}_{j=1}^{N}$ and the \textit{bright parameters} are
$\{a_{j}\}_{j=1}^{N}.$  FNNs generate their estimators conventionally
through solving the optimization problem
\begin{equation}\label{classical FNN training}
             \min_{(a_{j},\alpha_{j},b_{j})\in \mathbb{R}^{1}\times \mathbb R^{d}\times \mathbb{R}%
                ^{1}}\sum_{i=1}^{m}\left\vert \sum_{j=1}^{N}a_{j}\phi (\alpha_{j}\cdot
               x_{i}-b_{j})-y_{i}\right\vert ^{2}.
\end{equation}%
It is obvious that (\ref{classical FNN training}) does not distinguish  \textit{hidden parameters} and
\textit{bright parameters} and   is actually an OSL.

Our idea to design a TSL learning system based on FNNs   mainly stems from two interesting observations. On the
one hand, we observe in theoretical  literature \cite{Petrushev1999,Maiorov1999a} that to realize the optimal approximation capability,  the inner weights of an FNN can be restricted on the unit sphere embedded into the input space. This theoretical finding provides an intuition to design efficient learning schemes based on FNNs with shrinking the class of parameters. On the other hand,  the existing LtRaHP schemes \cite{Huang2006,Jaeger2004,Liu2013,Pao1989}  shows that the
uniform distribution for \textit{hidden parameters} is usually effective. This prompts
us to assign the \textit{hidden parameters} as uniform as possible.
An extreme assignment  is to deterministically select
 the
\textit{hidden parameters} as the equally spaced points (ESPs),
rather than   the random  sketching. Combining these two observations, it is reasonable
to generate inner weights as ESPs on the unit sphere and thresholds as ESPs on some interval.

The problem is, of course, can ESPs on the sphere be practically constructed?
This problem, known as the
Tamme's problem or the hard sphere problem \cite{Rogers1964}, is a
 well known and long-standing open question. This perhaps
explains why only LtRaHP has been widely utilized in TSL up to now,
even though several authors have already conjectured that LtDaHP may
outperform its random counterpart \cite{Ham2011}. However,
 due to the non-boundary property of the
sphere,  the Tamme's problem is the limiting
case of another famous problem: The minimal Riesz energy
configuration problem \cite{Saff1997}. The latter problem,
  listed as the $7$-th problem of Smale's ``problems for this century'' \cite{Smale1998},
can be approximately solved by using several methods, such as the
equal-area partition \cite{Saff1997} and  recursive zonal sphere
partitioning \cite{Leopardi2007}.
Thus, the \textit{hidden parameters} of FNNs can be
selected by appropriately combining the minimal Riesz energy points
 on the sphere with
the equally spaced points in an interval.

\subsection{Minimal Riesz energy points on the sphere}
 Let $  \mathbb{S}^{d-1}$ denote the unit sphere in
the $d$-dimensional Euclidean space $\mathbb{R}^{d}$, and $\Xi
_{n}:=\{\xi _{1},\dots ,\xi _{n}\}$ be a collection of $n$ distinct
points (a configuration) on $\mathbb{S}^{d-1}$. The Riesz $\tau
$-energy $(\tau \geq 0)$ associated with $\Xi _{n}$, denoted by
$A_{\tau }(\Xi _{n})$, is defined by \cite{Hardin2004}
$$
           A_{\tau }(\Xi _{n}):=\left\{
          \begin{array}{cc}
           \sum_{i\neq j}|x_{i}-x_{j}|^{-\tau }, & if\ \tau >0, \\
            \sum_{i\neq j}-\log |x_{i}-x_{j}|, & if\ \tau =0.%
           \end{array}%
           \right.
$$
Here $|\cdot |$ is the Euclidean norm. We use $ \mathcal{E}_{\tau
}(\mathbb S^{d-1},n)$ to denote the $n$-point minimal $\tau $-energy
over $\mathbb{S}^{d-1}$, that is,
\begin{equation}
\mathcal{E}_{\tau }(\mathbb{S}^{d-1},n):=\min_{\Xi _{n}\in \mathbb{S}%
^{d-1}}A_{\tau }(\Xi _{n}),  \label{minimal energy points set}
\end{equation}%
where the minimization is taken over all $n$-point configurations of $%
\mathbb{S}^{d-1}$. If $\Xi _{n}^{\ast }\subset \mathbb{S}^{d-1}$ is
a minimzer of (\ref{minimal energy points set}), i.e.,
$$
            A_{\tau }(\Xi _{n}^{\ast })=\mathcal{E}_{\tau
              }(\mathbb{S}^{d-1},n),
$$
then $\Xi _{n}^{\ast }$ is called a minimal $\tau $-energy configuration of $%
\mathbb{S}^{d-1}$, and the points in $\Xi _{n}^{\ast }$ are called
the minimal $\tau $-energy points.

The elegant work in \cite{Kuijlaars2007} showed
that the minimal $\tau $-energy points of $\mathbb{S}^{d-1} $ are an
effective approximation of the equally spaced points (ESPs) on the
sphere whenever $\tau\geq d-1$. Thus, one can use the minimal $\tau
$-energy points to substitute ESPs in applications.
As formulated as the Smale's $7th$ problem \cite{Smale1998}, generating minimal $\tau $%
-energy configurations and  minimal $\tau $-energy
points on $\mathbb{S}^{d-1}$ has triggered enormous research
activities
 \cite{Hardin2004,Leopardi2007,Saff1997} in the past thirty years.

Up till now,
there have been several well established approaches to approximately solve the
minimal $%
\tau$-energy configuration problems
\cite{Krantz1993,Leopardi2007,Saff1997}, among which two widely used
procedures are   Saff et al.'s equal-area partitioning
\cite{Saff1997} and  Leopardi's recursive zonal sphere partitioning
procedure \cite{Leopardi2007}. Both of them have been justified to
be able
to approximately generate the  minimal $\tau$-energy  points of $\mathbb{S}%
^{d-1} $ for a certain $\tau$ with a ``cheap'' computational cost,
more precisely, with an $\mathcal{O}(n\log n)$ asymptotic time
complexity \cite{Leopardi2007}.

\subsection{The LtDaHP Scheme}

Let ${\bf z}=\{(x_i,y_i)\}_{i=1}^m$ be the set of samples with
$x_i\in X$ and $y_i\in Y$. Without loss of generality, we assume the
input space $X=\mathbb{B}^{d}$ and the output space $Y\subseteq
\lbrack -M,M]$, where $\mathbb{B}^{d}$ is the unit ball in
$\mathbb{R}^{d}$ and $M>0$. Our idea is to solve an FNN-learning
problem by the TSL approach which deterministically assigns the
\textit{hidden parameters}  at the first stage, and solves a linear
least-square  problem at the second stage. In particular,   we
propose to deterministically assign   inner weights to be minimal
Riesz $\tau $-energy points of $\mathbb{S}^{d-1},$ and  thresholds
to be   ESPs in the interval $[-1/2,1/2].$ Consequently, our
suggested FNN-realization of   LtDaHP can be formalized as follows:
\bigskip

\textbf{LtDaHP Scheme:}\textit{\ Given the training samples }$%
{\bf z}=(x_{i},y_{i})_{i=1}^{m}$\textit{, the nonlinear function
}$\phi $\textit{\ and a splitting }$N=n\ell,$\textit{\ we generate
the LtDaHP estimator via the following two stages:}

\textit{Stage 1:} \textit{Take the inner weights }$\{\alpha
_{j}\}_{j=1}^{n}$\textit{\ to be   minimal} \textit{Riesz }$\tau $%
\textit{-energy points of }$\mathbb{S}^{d-1}$ \textit{with
$\tau\geq d-1$}, \textit{\ and }$\{b_{k}\}_{k=1}^{\ell}$\textit{\ to be  ESPs in the interval }$[-1/2,1/2],$\textit{\ that is, }%
$$
             b_{k}=-\frac12+\frac{k}{\ell},\ k= 1,2,\dots,\ell.
$$
 \textit{We then obtain a parameterized hypothesis space }
\begin{equation}\label{LDHP}
             \mathcal{H}_{\ell,n,\phi }:=\left\{
             \sum_{j=1}^{n}\sum_{k=1}^{\ell}a_{jk}\phi (\alpha
                _{j}x-b_k):a_{jk}\in \mathbb{R}^{1}\right\} .
\end{equation}

\textit{Stage 2:} \textit{The LtDaHP estimator is defined by}
\begin{equation}\label{Estimator}
          f_{\mathbf{z},\ell,n,\phi }:=\arg \min_{f\in \mathcal{H}_{\ell,n,\phi }}\frac{1}{m}%
             \sum_{i=1}^{m}|f(x_{i})-y_{i}|^{2}.
\end{equation}

Classical neural network approximation literature
\cite{Chen1993,Hahm2004,Cao2009} shows that neural networks with
fixed inner weights are sufficient to approximate univariate
functions. We adopt  this approach in our construction (\ref{LDHP})
by using $\sum_{k=1}^\ell a_k\phi(t-b_k)$ to approximate univariate
functions.  Then, we use an approach from \cite{Petrushev1999} to
extend univariate approximation to multivariate approximation (see
Section C in Appendix for detailed construction), which requires
$n\sim \ell^d$ different inner weights on the sphere and obtained an
FNN with good approximation property formed as (\ref{LDHP}). It
should be mentioned that $\ell$ in the splitting is the main
parameter   to control the approximation accuracy and $n$ depends on
$\ell$ is required for a dimensional extension. Based on the
splitting,  each inner weight shares $\ell$ same thresholds in
constructing  FNNs, which is different from the classical FNNs in
(\ref{classical FNN training}). The structures of functions in
$\mathcal{H}_{\ell,n,\phi }$ is shown in the following Figure 1.
 \begin{figure}[!t]
 \centering
 \includegraphics*[scale=0.30]{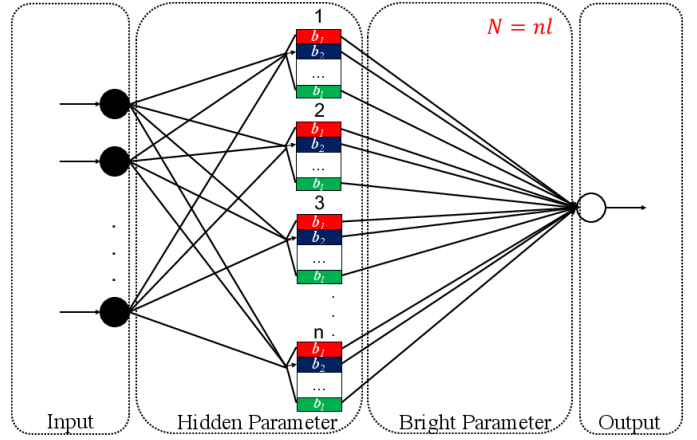}
 \hfill \caption{Topological structures of FNN-realization of LtDaHP.}
 \end{figure}

Based on the deterministic assignment of hidden parameters,  LtDaHP
then transforms a nonlinear optimization problem (\ref{classical FNN
training}) into a linear one (\ref{Estimator}), which reduces
heavily the computational burden. It should be mentioned that for
ESNs, there is another approach to deterministically construct
hidden parameters \cite{Rodan2012}. However, ESNs focus on training
recurrent neural networks rather than the  standard FNN studied in
this paper.

\section{Theoretical assessment}

In this section we study   theoretical behaviors of LtDaHP.  After
reviewing some basic notations of learning theory \cite{Cucker2007},
we prove that the FNN-realization of LtDaHP  provides an almost
optimal generalization error bound as long as  the regression
function is smooth.

\subsection{Statistical Learning Theory}

Suppose that $\mathbf{z}=(x_{i},y_{i})_{i=1}^{m}$ are drawn
independently and identically from $Z:=X\times Y$ according to an
unknown probability distribution $\rho $ which admits the
decomposition
$$
                 \rho (x,y)=\rho _{X}(x)\rho (y|x).
$$
Assume that $f:X\rightarrow Y$ is a function that characterizes the
correspondence between the input and output, as induced by $\rho $.
A natural measurement of the error incurred by using $f$ of this
purpose is the generalization error, defined by
$$
             \mathcal{E}(f):=\int_{Z}(f(x)-y)^{2}d\rho ,
$$
which is minimized by the regression function \cite[Chap.1]{Cucker2007}
$$
                f_{\rho }(x):=\int_{Y}yd\rho (y|x).
$$
We do not know this ideal minimizer $f_{\rho }$  since $\rho $ is
unknown, but we have access to random examples ${\bf z}$ from
$X\times Y$ sampled according to $\rho $.

Let $L_{\rho _{_{X}}}^{2}$ be the Hilbert space of $\rho
_{X}$-square-integrable functions on $X$, with norm $\Vert \cdot
\Vert _{\rho }.$ It is known that, for every $f\in L_{\rho
_{X}}^{2}$, there holds \cite[Chap.1]{Cucker2007}
\begin{equation}\label{equality}
          \mathcal{E}(f)-\mathcal{E}(f_{\rho })=\Vert f-f_{\rho }\Vert _{\rho
           }^{2}.
\end{equation}%
So, the goal of learning is to find a best approximation of the
regression function $f_{\rho }$.
%

If we have a specific estimator $f_{\mathbf{z}}$ of $f_{\rho }$ in
hand, the error $\mathcal{E}(f_{\mathbf{z}})-\mathcal{E}(f_{\rho })$
clearly depends on $\mathbf{z}$ and therefore has a stochastic
nature. As a result, it is
impossible to say anything about $\mathcal{E}(f_{\mathbf{z}})-\mathcal{E}%
(f_{\rho })$ in general for a fixed $\mathbf{z}$. Instead, we can
look at its behavior in probability as measured by the following
expected error
$$
              \mathbf E_{\rho ^{m}}(\Vert f_{\mathbf{z}}-f_{\rho }\Vert _{\rho
              }^{2}):=\int_{Z^{m}}\Vert f_{\mathbf{z}}-f_{\rho }\Vert_\rho ^{2}d\rho
               ^{m},
$$
where the expectation is taken over all realizations $\mathbf{z}$
obtained for a fixed $m$, and $\rho ^{m}$ is the $m$ fold tensor
product of $\rho $.

It is known \cite{Zhou2006} that whenever $y\in [-M,M],$
using the truncation operator $\pi _{M}$ can reduce  the
generalization error of $f_{\bf z}$ without adding extra computation. Thus, instead of using $f_{\mathbf{z},\ell,n,\phi }$ defined by (\ref{Estimator}), we  take $\pi _{M}f_{%
\mathbf{z},\ell,n,\phi }(x)$ as the   LtDaHP\ estimator, where $\pi _{M}f(x):=\mbox{sign}(f(x))\min\{M,|f(x)|\}$
 is the truncation operator on $f(x)$.

\subsection{An Almost Optimal Generalization Bound}

In general, it is impossible to get a nontrivial generalization error
bound   of a learning algorithm without knowing any information on $\rho $ \cite[Thm.3.1]{Gyorfi2002}. So, some types of a-priori information
of the regression function $f_{\rho }$ have to
be imposed. Let $\mathbb{N}$ be   the set of positive integers and $\mathbf{k}=(k_{1},k_{2},\dots ,k_{d})$ with each $%
k_{i}\in
\mathbb{N}
.$ The ${\bf k}$-th order derivative of a function $f$ is defined by
$$
         D^{\mathbf{k}}f(x):=\frac{\partial ^{|\mathbf{k}|}f}{\partial
            ^{k_{1}}x^{(1)}\cdots \partial ^{k_{d}}x^{(d)}},
$$
where $|\mathbf{k}|:=k_{1}+\cdots +k_{d} $ and $x=(x^{(1)},\dots,x^{(d)})$. The classical Sobolev
class is then defined for any $r\in
\mathbb{N}
$ by
$$
           W_{p}^{r}:=W_{p}^{r}(\mathbb{B}^{d}):=\left\{ f:\mathbb{B}^{d}\rightarrow
           \mathbb{R}^{1}:\max_{0\leq |\mathbf{k}|\leq r}\Vert D^{\mathbf{k}}f\Vert
           _{p}<\infty \right\} .
$$
Let $J$ be the identity mapping
$$
               L^2(\mathbb B^d) ~~ {\stackrel{J}{\longrightarrow}}~~ L^2_{\rho_X}.
$$
and $D_{\rho _{X}}=$ $\Vert J\Vert .$ $D_{\rho _{X}}$ is called the
distortion of $\rho _{X}$, which measures how much $\rho _{X}$
distorts the Lebesgue measure. We assume that the distribution $\rho
$ satisfies $D_{\rho _{X}}<\infty $ and $f_{\rho }\in W_{2}^{r}$,
which is standard and  utilized in vast literature
\cite{Gyorfi2002,Cucker2007,Maiorov2006a,Shi2011,Lin2013b,Lin2018}.

Since the generalization capability of LtDaHP depends also on the
  activation function $\phi ,$ certain restrictions on $\phi
$ should be  imposed.  We   say that $\phi $ is a sigmoid function,
if $\phi $ satisfies
$$
           \lim_{t\rightarrow \infty }\phi (t)=1,\quad \lim_{t\rightarrow -\infty }\phi
           (t)=0.
$$
 By   definition,
for any sigmoid  function $\phi ,$ there   exists a  positive
constant $L$ such that
\begin{equation}\label{definition
L}
         \left\{\begin{array}{cc} |\phi (u)-1|<m^{-\frac{2}{2r+d}}, & \mbox{if}\quad u\geq L,\\
               |\phi (u)|<m^{-\frac{2}{2r+d}}, & \mbox{if}\quad u\leq
               -L. \end{array}\right.
\end{equation}
Define
$$
                   \phi_{K}(t):=\phi (Kt),
$$
for any
\begin{equation}\label{choose K}
              K\geq \ell L,
\end{equation}%
where $\ell$ is the number of different thresholds in the LtDaHP scheme.
We further suppose that for arbitrary closed set $A$ in $\mathbb{R}^{1},$ $%
\phi $ is square integrable, which is denoted  by $\phi \in L_{Loc}^{2}(%
\mathbb{R}^{1}).$

Our main result  is the following Theorem
\ref{THEOREM1}, which shows that $f_{\mathbf{z},\ell,n,\phi _{K}},$ the
defined LtDaHP estimator (\ref{Estimator}), can achieve
an almost optimallearning rate.

\begin{theorem}\label{THEOREM1}
Let $d\geq 2$. Assume $0<r\leq \frac{d+1}{2}$, $\phi \in
L_{Loc}^{2}(\mathbb{R}^{1})$ is a bounded sigmoid  function and
$f_{\mathbf{z},\ell,n,\phi _{K}}$ is the LtDaHP\ estimator defined by
(\ref{Estimator}). If $f_{\rho }\in W_{2}^{r}$,  $\ell=\left[
m^{\frac{1}{d+2r}}\right] $, $n\sim l^{d-1}$
and $K$ satisfies (\ref{choose K}), then there exist positive constants $%
C_{1}$ and $C_{2}$, depending only on $d,r,M$ and $\phi $, such
that,
\begin{equation}\label{theorem1}
               C_{1}m^{-\frac{2r}{d+2r}}\leq \sup_{f_{\rho }\in W_{2}^{r}}\mathbf E_{\rho
                 ^{m}}(\Vert f_{\rho }-\pi _{M}f_{\mathbf{z},\ell,n,\phi _{K}}\Vert
                _{\rho }^{2})\leq C_{2}D_{\rho _{X}}^{2}m^{-\frac{2r}{d+2r}}\log m.
\end{equation}
\end{theorem}

The proof of Theorem \ref{THEOREM1} will be presented in Appendix. Some immediate remarks, to
explain this result, are as follows.

\subsection{Remarks}

\subsubsection{On optimality of generalization error}

We see that modulo the logarithmic factor $\log m$, the established
learning rate (\ref{theorem1}) is optimal  in a minmax sense. That
is, up to a logarithmic factor, the upper and lower bounds
 of the learning rate are asymptotically identical.
We further show that this learning rate is also almost
optimal among all   learning schemes. Let
$\mathcal{J}(W_{2}^{r})$ be the class of all Borel measures $\rho $
 satisfying $f_{\rho }\in W_{2}^{r}$ and $D_{\rho _{X}}<\infty
$. We enter into a competition over all estimators $\mathcal
A_{m}:\mathbf{z}\rightarrow f_{\mathbf{z}}$ and   define
$$
              e_{m}(W_{2}^{r}):=\inf_{\mathcal A_{m}}\sup_{\rho \in
             \mathcal{J}(W_{2}^{r})}\mathbf E_{\rho ^{m}}(\Vert f_{\rho
              }-f_{\mathbf{z}}\Vert _{\rho }^{2}).
$$
Then, $e_{m}(W_{2}^{r})$ quantitatively measures the quality of $f_{\mathbf{z}%
}$ and it was shown in \cite[Chap. 3]{Gyorfi2002}  that
\begin{equation}
e_{m}(W_{2}^{r})\geq Cm^{-\frac{2r}{2r+d}},\ m=1,2,\dots ,
\label{baseline}
\end{equation}%
where $C$ is a constant depending only on $M$, $d$ and $r.$ (\ref%
{baseline}) shows that if $f_{\rho }\in W_{2}^{r}$ and $D_{\rho
_{X}}<\infty ,$  learning rates of all learning strategies based on
$m$ samples cannot be faster than
$\mathcal{O(}m^{-\frac{2r}{2r+d}}).$
Consequently, the  learning rate established in (\ref%
{theorem1}) is  almost optimal among all learning schemes.

In this sense, Theorem \ref{THEOREM1}  says that  even when the \textit{hidden
parameters} are not trained and just preassigned deterministically,  LtDaHP
does not degrade the generalization capability of FNNs which train
  \textit{hidden and bright parameters} together by some OSL scheme.
It is noted that a similar almost optimal learning rate has also
been proved in \cite{Liu2013} for a typical scheme of LtRaHP (ELM):
\begin{equation}\label{ELM}
       C_{1}m^{-\frac{2r}{d+2r}}\leq \sup_{f_{\rho }\in W_{2}^{r}}\mathbf E_{\mu
        }\mathbf E_{\rho ^{m}}(\Vert f_{\rho }-\pi _{M}f_{LRHP}\Vert _{\rho }^{2})\leq C_{2}D_{\rho
          _{X}}^{2}m^{-\frac{2r}{d+2r}}\log m,
\end{equation}%
in which the expectation $\mathbf E_{\mu }$ is taken over all
possible random assignments of   \textit{hidden parameters}. We
refer the readers to \cite{Liu2013} for detailed definitions of
$f_{LRHP}$ and $\mu$. There is an additional expectation term that
brings the uncertainty problem of LtRaHP. Comparing (\ref{ELM}) with
(\ref{theorem1}),
we can see that the LtDaHP   dismisses the $%
\mathbf E_{\mu }$-expectation term. Furthermore, we notice that
LtRaHP, as shown in \cite{Lin2014}, may break the almost optimal
generalization error for certain specific   activation
functions even in the $\mathbf E_{\mu }$-expectation sense. Theorem
\ref{THEOREM1} thus implies that the LtDaHP improves on LtRaHP not
only in circumventing  the uncertainty problem, but also guaranteeing
the generalization capability further.

\subsubsection{On how to specify the activation  function $\protect%
\phi $}

In Theorem \ref{THEOREM1}, three conditions have been imposed on
the activation function $\phi :\ $\textit{(i)}
$\phi_{K}(t)=\phi (Kt)$ with $K\geq \ell L$, where $L$ is defined by
(\ref{definition L}), \textit{\ (ii)} $\phi $ is a bounded sigmoid function, and $%
(iii)$ $\phi \in L_{Loc}^{2}(\mathbb{R}^{1}).$ The
conditions\textit{\ (ii) }and \textit{(iii)} are clearly mild,
say, both the widely applied heaviside function $\sigma_{H}$ and
logistic function $\sigma_{L}$   satisfy the assumptions, where
$$
            \sigma_{H}(t):=\left\{
            \begin{array}{cc}
          1, & t\geq 0 \\
                0, & t<0%
           \end{array}%
                \right.
$$
and
$$
               \sigma_{L}(t)=\frac{1}{1+e^{-t}}.
$$
The most crucial assumption  is \textit{(i), }i.e., $K$ should be
  carefully chosen. It is observed that  $K$ in
(\ref{choose K}) is
with respective to $\ell$ and $L$, while $L$ depends merely on $m$ and $\phi .$ So $%
K $ can be specified when $\phi $ is given. For example, if $\phi $
is the logistic function, then $K$ can be selected as any positive
numbers satisfying $K\geq \ell\log (\ell^{2}-1)$.

The problem is that there are infinite many choices of such $K$. How
to specify the best $K$ thus becomes a practical issue. According to
\cite[Thm. 2.4]{Maiorov2006}, the complexity of
$\mathcal{H}_{{\ell,n,\phi _{K}}}$ monotonously increases with
respect to $K$. Due to the well known bias and variance trade-off
principle, we then recommend to choose $K=\ell L$ in practice. For
example, when the logistic
function is utilized, we may take $K=\ell\log (\ell^{2}-1)$ if $\ell\geq 2,$ and $%
K= \log 2$ if $\ell=1$.

\subsubsection{On almost optimality of the number of hidden neurons}

Theorem \ref{THEOREM1} has presented the number of hidden neurons to
be $n\ell =\mathcal O(m^{\frac{d}{2r+d}}).$
We observe from (\ref{LDHP}) that $\mathcal{H}_{\ell,n,\phi _{K}}$ is an $n\ell$%
-dimensional linear space. Hence, according to the well known linear
width theory \cite{Pinkus1985}, for $f_{\rho }\in W_{2}^{r}$,
$n\ell$ must be not smaller than $\mathcal O(m^{\frac{d}{2r+d}})$ if
one wants to achieve an approximation error of
$\mathcal{O}(m^{\frac{-2r}{2r+d}}).$ This means that the number of
hidden neurons required in Theorem \ref{THEOREM1} cannot be reduced.

At the first glance, there are two parameters, $\ell$ and $n$, that
need to be specified, which is more complicated than that in OSL
\cite{Maiorov2006a} and LtRaHP \cite{Liu2013}. In fact, there is
only an essential parameter since   $n$ and $\ell$ have a relation
$n\sim \ell^{d-1}$.   If the smoothness information of $f_{\rho }$
is known, one can directly take $n\sim \ell^{d-1}=\left[
m^{\frac{1}{d+2r}}\right]^{d-1} $ as that in Theorem \ref{THEOREM1}.
However, it is usually infeasible since we  do not know the concrete
value of $r$, when faced with real world applications. Instead, we
turn to some model-selection strategy such as the \textquotedblleft
cross-validation\textquotedblright\ approach
\cite[Chap.8]{Gyorfi2002} to determine $\ell$ and $N$.

\subsubsection{On why LtDaHP works}

 It is known \cite[Chap.1]{Cucker2007} that a satisfactory
generalization capability  of a learning scheme can only be resulted
from an appropriate trade-off   between the approximation
capability and capacity of the hypothesis space. We  use this principle to explain the success of LtDaHP.

Given a hypothesis space $\mathcal{F}$ and a function $f,$ the
approximation capability of $\mathcal{F}$ can be measured by its
best
approximation error:%
$
                dist(f,\mathcal{F}):=\inf_{g\in \mathcal{F}}\Vert f-g\Vert _{2}
$
while the capacity of $\mathcal{F}$ can be measured by its
pseudo-dimension \cite{Maiorov2006}, denoted by $Pdim(\mathcal F)$.
We  compare the approximation
capabilities and
capacities of the hypothesis spaces of FNNs and LtDaHP. The hypothesis space $%
\mathcal {N}_{N,K}^{\ast }$ of FNNs is the family of functions of the form%
$$
               N_{N,K}(x):=\sum_{j=1}^{N}c_{j}\phi (\alpha_{j}\cdot x-b_{j})
$$
where $\|\alpha_{j}\|\leq K$ and $|b_{j}|\leq K$. In \cite{Maiorov2006} and \cite{Lin2018a},
 it was shown that for some positive constant $c$ there
holds
\begin{equation}\label{Classical pdim}
            P dim(\mathcal{N}_{N,K}^{\ast })\leq cd^{2}N\log (KN),
\end{equation}%
if $\phi $ is the logistic function. Furthermore, in
\cite{Maiorov2006a}, it was verified the approximation capability of $\mathcal N_{N,K}^{\ast
}$ satisfies
\begin{equation}\label{classical approximation error}
        dist(f,\mathcal{N}_{N,K}^{\ast })\leq CN^{-r/d}.
\end{equation}%
provided $f_{\rho }\in
W_{2}^{r}.$

In comparison, we find in  \cite{Mendelson2003}  that
\begin{equation}\label{LDHP pdim}
             P dim(\mathcal{H}_{{\ell ,n,\phi _{K}}})=N
\end{equation}%
and, similarly, we can prove  in Lemma \ref{Lemma:APPROXIMATION ERROR} that
\begin{equation}\label{LDHP approximation error}
          dist(f,\mathcal{H}_{\ell,n,\phi _{K}})\leq c(n\ell)^{-r/d}=cN^{-r/d},
\end{equation}%
as long as $K$ satisfies (\ref{choose K}). Comparing (\ref{Classical
pdim}) and (\ref{classical approximation error}) with (\ref{LDHP
pdim}) and (\ref{LDHP approximation error}), we thus conclude that for
appropriately tuned $K$,
both the approximation capabilities and capacities of $\mathcal{N}%
_{N,K}^{\ast }$ and $\mathcal{H}_{l,n,\phi _{K}}$ are almost the
same. This shows the reason why the LtDaHP scheme  performs at least not worse
than the conventional FNNs.

\section{Experimental Studies}

In this section, we present both toy simulations and real world data
experiments to assess the performance of LtDaHP as compared with support vector regression (SVR), Gaussian process regression (GPR), and a typical LtRaHP scheme (ELM), where the learning algorithm is the same as LtDaHP, except that the inner weights
and thresholds are randomly sampled according to the uniform
distribution. In our experiments, the minimal
R\textit{iesz }energy points were approximately generated by the
recursive zonal sphere partitioning \cite{Leopardi2007}
using the EQSP tool box\footnote{%
http://www.mathworks.com/matlabcentral/fileexchange/13356-eqsp-recursive-zonal-sphere-partitioning-toolbox%
}. SVR and GPR were realized by
the Matlab functions fitrsvm and fitrgp(with subsampling 1000 atoms), respectively. For fair of comparisons, we applied the
10-fold cross-validation method \cite[Chap.8]{Gyorfi2002} to select all
parameters (more specifically, to select three parameters in
SVR, the width of Gaussian kernel, the regularization parameter and the epsilon-insensitive band,
and  two parameters in LtRaHP and LtDaHP, the number of hidden
neurons for both methods, $\ell$ for LtDaHP, and $K$ for LtRaHP), and the time for parameter tuning was included in recording the training time.

All the simulations and
experiments were conducted in Matlab R2017b on a workstation
with 64Gb RAM and E5-2667 v2 3.30GHz CPU.

\subsection{Toy Simulations}

This series of simulations were designed to support the correctness
of Theorem \ref{THEOREM1} and compare the learning performance
among LtDaHP, LtRaHP, GPR, and SVR. For this
purpose, the regression function $f_{\rho }$ is supposed to be known and given by%
$$
f_{\rho }(x)=(1-\Vert x\Vert _{2})_{+}^{4}(4\Vert x\Vert _{2}+1),
x\in \lbrack -1,1]^{3}
$$
where $a_{+}=\max \{a,0\}.$ Direct computation shows $f_{\rho
}\in W_{2}^{2}$ and $f_{\rho }\notin W_{2}^{3}$. We generated the
training sample set ${\bf z}=\{(x_{i},y_{i})\}_{i=1}^{m}$ with
variable data size through independently and randomly sampling
$x_{i}$ from $[-1,1]^{3}$ according to the
uniform distribution, and $%
y_{i}=f_{\rho }(x_{i})+\epsilon $ with $\varepsilon \sim N(0,0.1)$
being the white noise. The learning performance of the algorithms
were  tested by
applying the resultant estimators to the test set ${\bf z}_{test}=%
\{(x_{i}^{(t)},y_{i}^{(t)})\}_{i=1}^{1000}$ which was generated
similarly
to ${\bf z}$  with a difference that  $%
y_{i}^{(t)}=f_{\rho }(x_{i}^{(t)}).$ In simulations, we took
 $\phi
(t)=\frac{1}{1+e^{-t}},$ and implemented the
LtDaHP and LtRaHP with $\phi_{K}(t)=\phi (Kt),$ where%
$$
        K=\left\{
           \begin{array}{cc}
           \ell\log (\ell^{2}-1), & \ell>1 \\
            \log 2, & \ell=1%
           \end{array}%
                   \right.
$$
for LtDaHP as suggested in the subsection III.C. The $\ell$ for
LtDaHP and $K$ for LtRaHP were tuned by 10-fold cross validation. In
addition, to avoid the risk of singularity, we implement the least
square (2) for LtDaHP and LtRaHP with a very small and fixed
regularization pamameter $\lambda=10^{-4}$.

\begin{figure}[!t]
\begin{minipage}[b]{.49\linewidth}
\centering
\includegraphics*[scale=0.31]{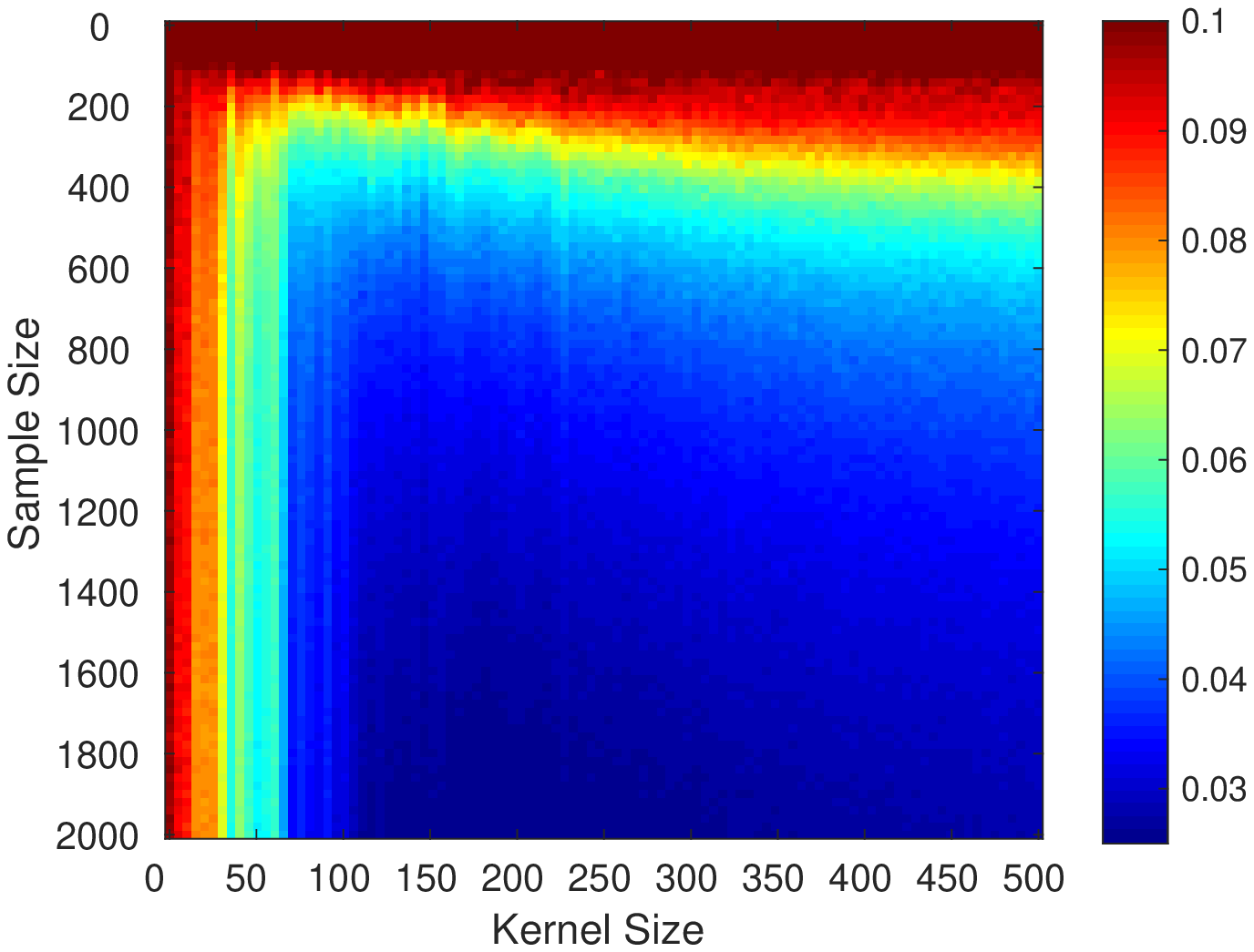}
\centerline{{\small (a) LtDaHP phase diagram}}
\end{minipage}
\hfill
\begin{minipage}[b]{.49\linewidth}
\centering
\includegraphics*[scale=0.31]{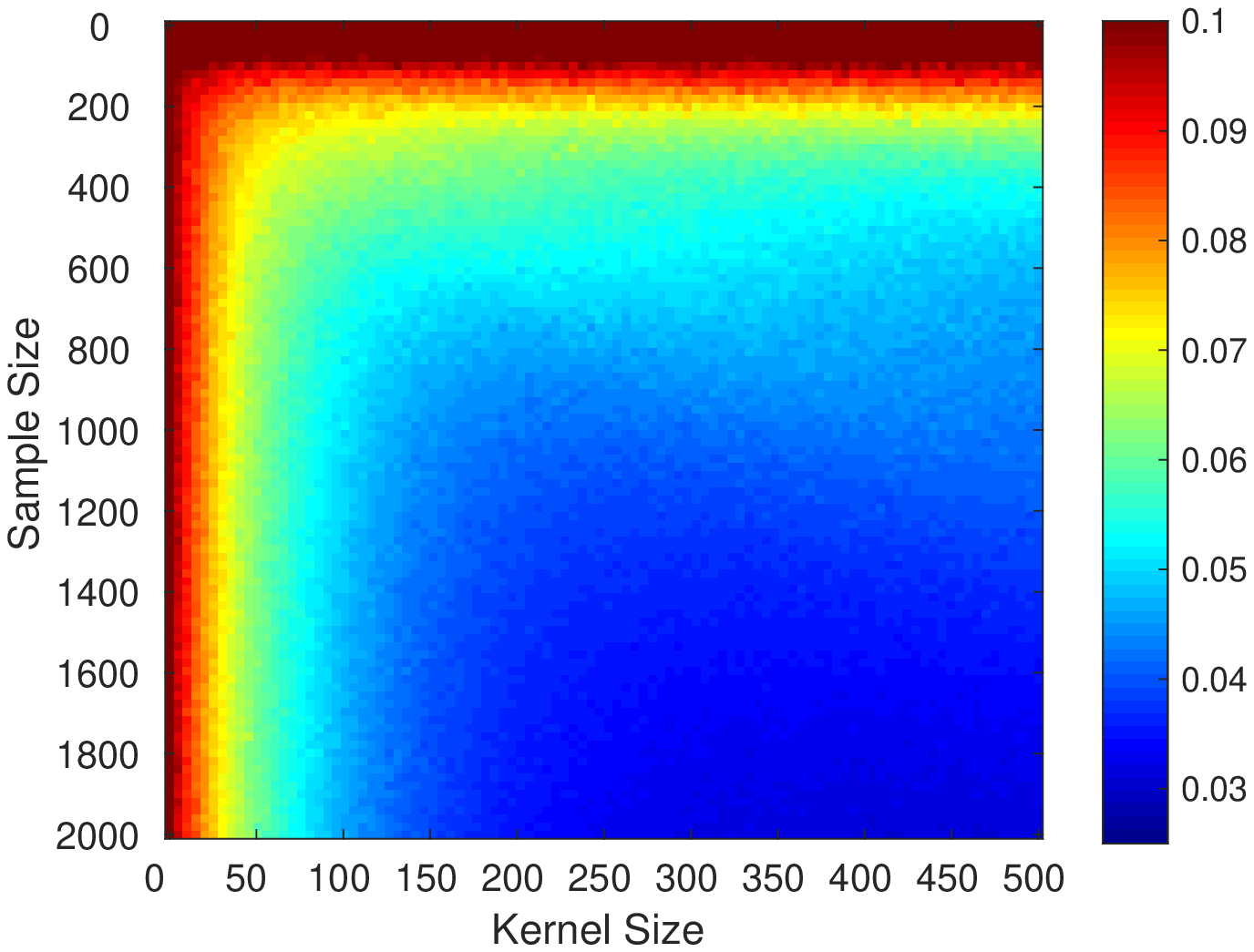}
\centerline{{\small (b) LtRaHP phase diagram}}
\end{minipage}
\hfill \caption{ The comparison of generalization capability between
LtDaHP and LtRaHP }
\label{fig:simu1}
\end{figure}

In the first simulation, to illustrate the difference between
LtDaHP and LtRaHP,  we conducted a phase
diagram study. To this end, the number of samples $m$ varied from
$20$ to $2000$ and the number of neurons $N$ ranged from $5$ to $500$. For each pair $(m,N)$, we
implemented 100 independent simulations with LtDaHP and LtRaHP. The
average rooted mean square errors (RMSE) were then recorded. We
plotted all
the simulation results in a figure, called \textit{the phase diagram,} with $%
x$-axis and $y$-axis being respectively the  number of neurons
 and the number of samples, and the colors from blue to red
corresponding to the RMSE values from small to large. The simulation
results are reported in Fig. \ref{fig:simu1}.

Fig. \ref{fig:simu1}(a)   shows that for suitable choice of  $N$, the LtDaHP
estimator maintains always very low RMSE, which coincides with Theorem \ref{THEOREM1}. Furthermore, comparing (a) with (b) in
Fig. \ref{fig:simu1} demonstrates several obvious differences between LtDaHP and
LtRaHP: \textit{(i) } the test errors of LtDaHP are much smaller than
those of LtRaHP. This can be observed not only for the best choice
of the number of neurons, but also for every fixed number of neurons
as well. This difference reveals  that as far as the generalization
capability is concerned, LtDaHP outperforms LtRaHP in this
example. \textit{\ (ii) } LtDaHP exhibits a somewhat tidy phase change
phenomenon: there is a clear range of $m$ and $N$ such that the
LtDaHP performs well. Similar   phase change phenomenon does not
appear in LtRaHP, as exhibited in Fig. \ref{fig:simu1}(b), due to   the
uncertainty. This difference implies that
LtDaHP is more robust to the specification of neuron number $N$ than
LtRaHP and
selecting an appropriate neuron number $N$ for LtDaHP is much
easier  than that for LtRaHP. All these differences show the
advantages of LtDaHP\ over LtRaHP.

\begin{figure*}[!t]
  \begin{minipage}[b]{.32\linewidth}
  \centering
  \includegraphics*[scale=0.4]{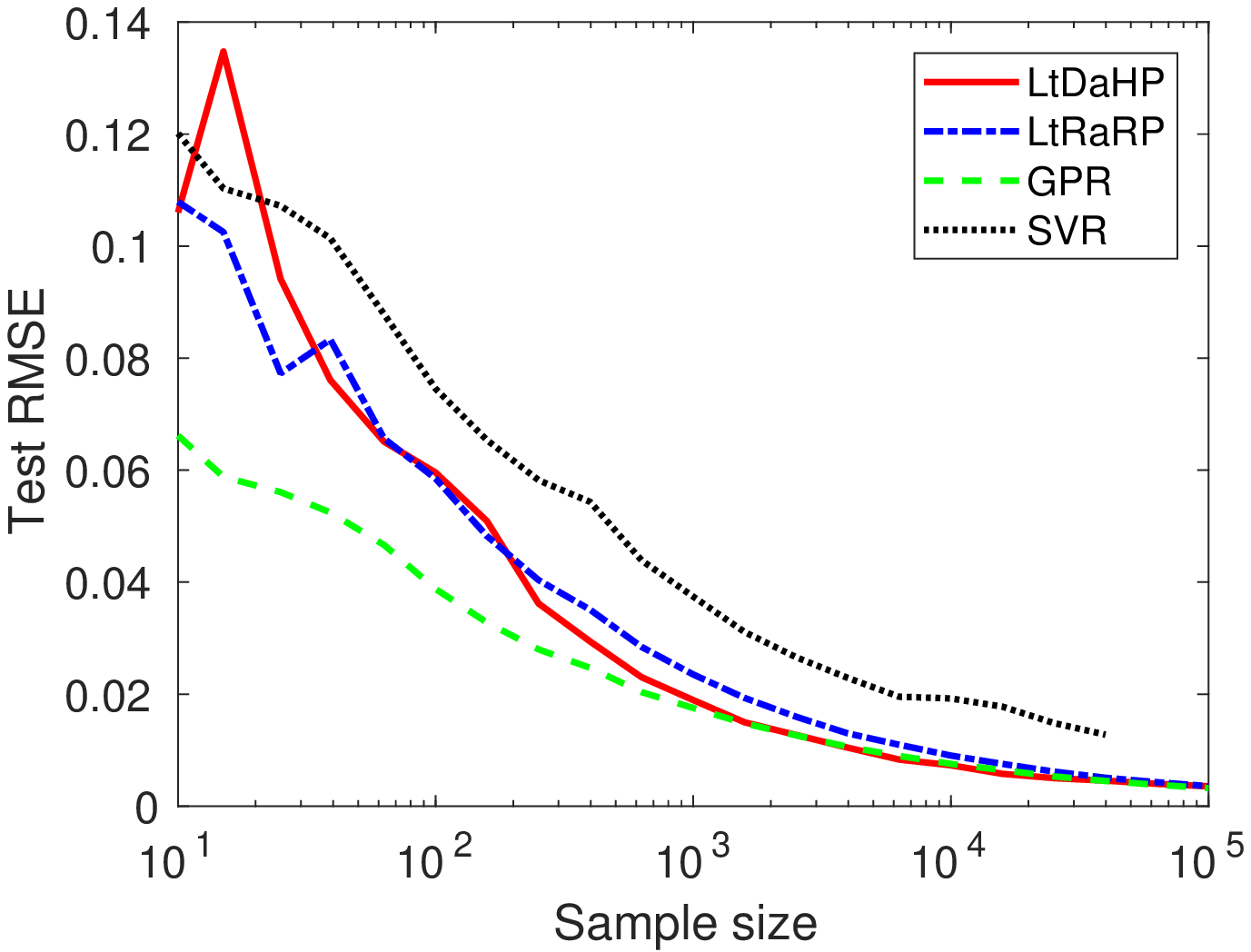}
  \centerline{{\small (a) Comparison of test error}}
  \end{minipage}
  \hfill
  \begin{minipage}[b]{.32\linewidth}
  \centering
  \includegraphics*[scale=0.4]{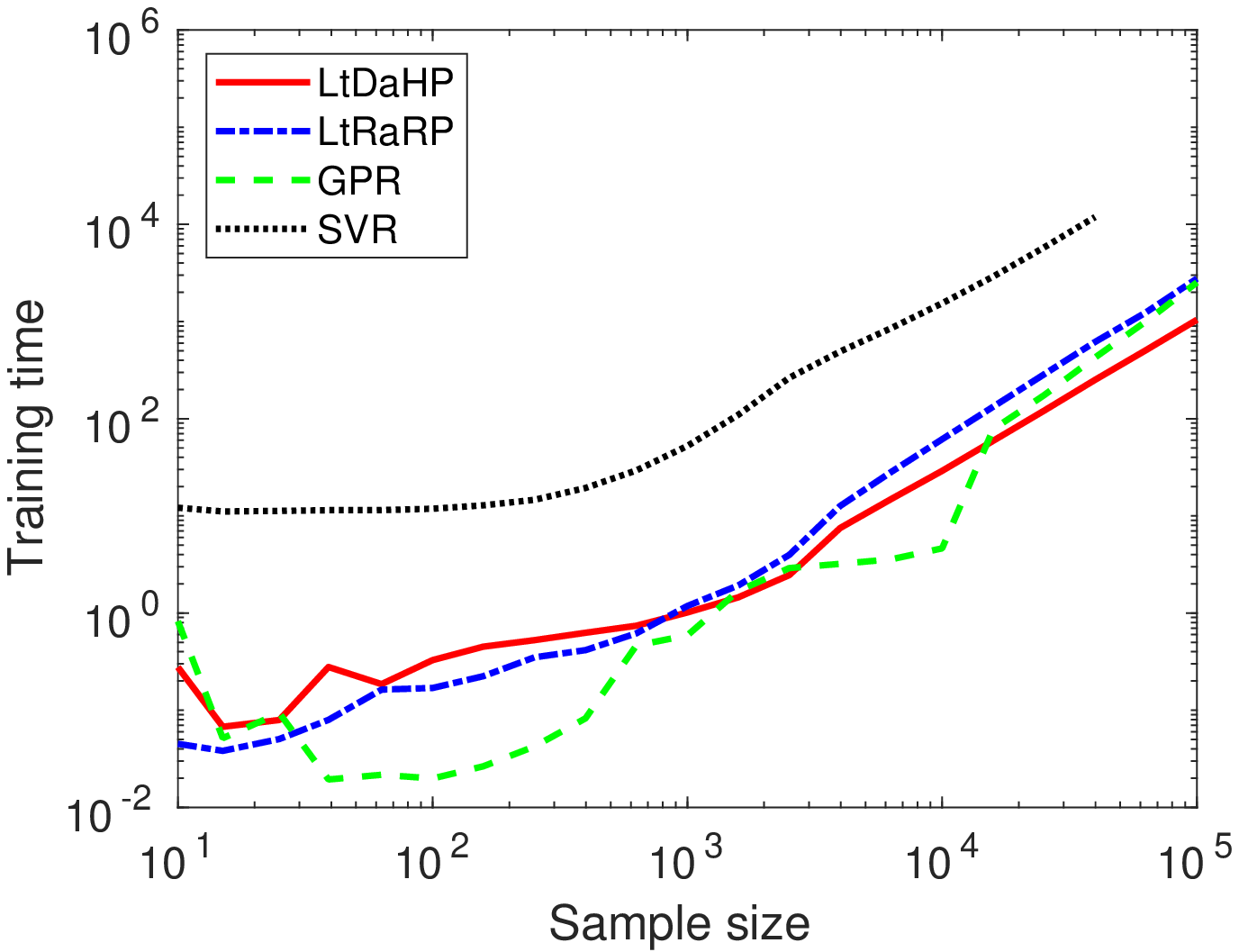}
  \centerline{{\small (b) Comparison of training time}}
  \end{minipage}
  \hfill
  \begin{minipage}[b]{.32\linewidth}
  \centering
  \includegraphics*[scale=0.4]{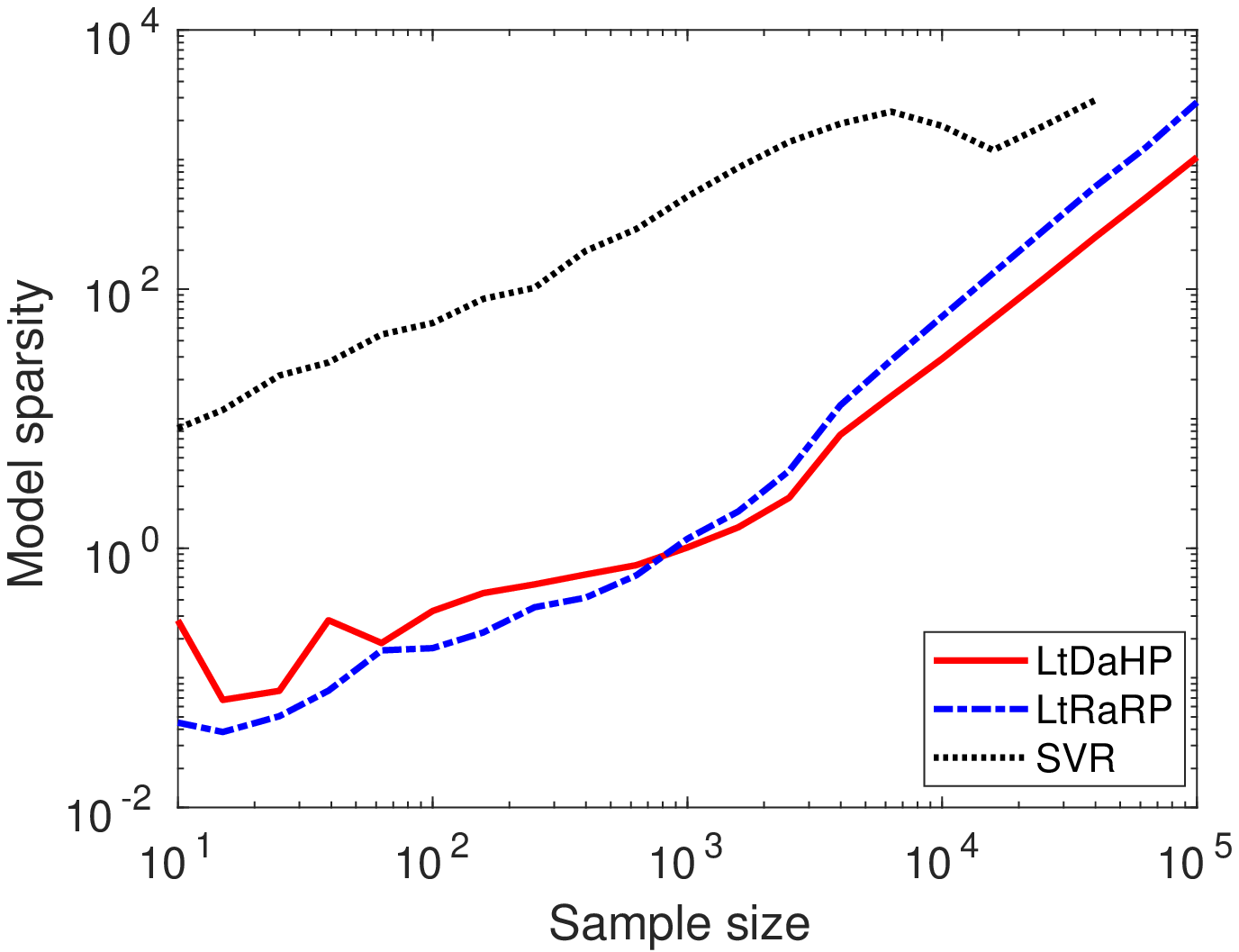}
  \centerline{{\small (c) Comparison of model sparsity}}
  \end{minipage}
  \hfill \caption{ The comparisons of test error, training time, and model sparsity
   among LtDaHP, LtRaHP, GPR, and SVR.}
   \label{fig:simu2}
  \end{figure*}

%
%

In the second simulation, we  studied the pros and cons of
 LtDaHP, LtRaHP,  GPR, and SVR.  We implemented these four algorithms independently
50 times and  calculated the average RMSEs. The obtained RMSEs,
as well as  the corresponding computational time and model sparsity, were
plotted as a function of the number of training samples
 in Fig. \ref{fig:simu2}.

From Fig. \ref{fig:simu2}(a), we can see that  GPR performs with the best
generalization capability, then LtDaHP, LtRaHP and finally SVR.
More specifically, we observe that the test errors of LtDaHP are very close to GPR when the sample size is over $1000$. But LtRaHP requires more samples to reach comparable performances. From Fig. \ref{fig:simu2}(b) we can see that LtDaHP and LtRaHP have constantly low training time. In this simulation, GPR has lower training time when samples are smaller than $10000$, but is less predictable due to more complex algorithm. In addition, SVR always takes more time for training, since there are 3 hyper-parameters to be tuned. Finally from Fig. \ref{fig:simu2}(c) we can see that LtDaHP and LtRaHP has much smaller model sparsity than SVR, which suggests less time in prediction.

All these simulations support the outperformance of LtDaHP and  the theoretical assertions made in the previous sections.

\subsection{Real World Benchmark Data Experiments}
\begin{table}[h]
\begin{center}
\begin{tabular}{|l|l|l|l|l|}
\hline Data sets & Training samples & Testing samplesr & Attributes\\
\hline Machine & 167 & 42 & 7 \\
\hline Yacht & 246 & 62& 6   \\
\hline Energy & 614 & 154 & 8   \\
\hline Stock & 760 & 190 & 9 \\
\hline Concrete &824 &206 &8 \\
\hline Bank8FM &3599 &900 &8 \\
\hline Delta\_ailerons & 5703 & 1426 & 5   \\
\hline Delta\_elevators &7613 & 1904 & 6   \\
\hline Elevator &13279 &3320 &18 \\
\hline Bike &13903 &3476 &17 \\
\hline
\end{tabular}%
\caption{Setting-up for data sets}
\end{center}
\end{table}

\begin{table*}[!th]
\begin{center}
\begin{tabular}{|c|c|c|c|c|c|c|c|c|c|c|c|}
\hline Data set & \multicolumn{4}{c|}{TestRMSE} &
\multicolumn{4}{c|}{TrainMT} & \multicolumn{3}{c|}{MSparsity} \\
\cline{2-12} & LtDaHP & LtRaHP &GPR & SVR  & LtDaHP & LtRaHP &GPR & SVR  & LtDaHP &
LtRaHP  & SVR  \\
\hline Machine & $0.084\pm 0.011$ & $0.081\pm 0.013$ & $0.082\pm 0.013$ &$0.105\pm 0.019$ &1.8 &0.3 &0.1 &2.6 &51.8 &45.4 &145.7 \\
\hline Yacht & $0.012\pm 0.007$ & $0.027\pm 0.009$& $0.012\pm 0.006$ &$0.049\pm 0.018$ &1.9 &0.5 &0.1 &3.0 &131.8 &123.6 &165  \\
\hline Energy & $0.015\pm 0.003$ & $0.053\pm 0.005$ & $0.012\pm 0.001$ & $0.019\pm 0.002$&3.5  &0.8 &1.6 &6.5 &195 &190 &439 \\
\hline Stock & $0.039\pm 0.003$ & $0.039\pm 0.003$ & $0.038\pm 0.003$ & $0.054\pm 0.009$ &3.3 &0.45 &0.24 &5.1 &76 & 106 & 377\\
\hline Concrete &$0.072\pm 0.005$ &$0.079\pm 0.006$ &$0.068\pm 0.006$ &$0.072\pm 0.007$ &4.1 &0.8 &0.9 &9.2 &204 &211 &606 \\
\hline Bank8FM &$0.040\pm 0.001$ &$0.040\pm 0.001$ &$0.037\pm 0.001$ &$0.039\pm 0.002$ &13 &9 &8 &222 &335 &439 &1938 \\
\hline Delta\_a & $0.038\pm 0.001$ & $0.038\pm 0.001$ &$0.039\pm 0.001$  &$0.040\pm 0.002$ &16 &14 &8 &222 &534 &497 &2617 \\
\hline Delta\_e &$0.053\pm 0.001$ & $0.053\pm 0.001$ & $0.053\pm 0.001$ &$0.053\pm 0.001$ &25 &22 &9 &357 &482 &534 &3564  \\
\hline Elevator &$0.052\pm 0.001$  &$0.052\pm 0.001$  &$0.052\pm 0.001$  &$0.054\pm 0.001$  &55 &51 &353 &703 &768 &730 &5455 \\
\hline Bike &$0.032\pm 0.001$  &$0.064\pm 0.002$  &$0.049\pm 0.001$  &$0.052\pm 0.002$  &57 &51 &389 &660 &792 &792 &3764 \\
\hline
\end{tabular}%
\caption{ Results of SVR, LtRaHP and LtDaHP when applied
to the 10 real world benchmark data sets}
\end{center}
\end{table*}

We further apply LtDaHP, LtRaHP, GPR, and SVR to a family of real world
benchmark data sets. We include 10 problems
covering different fields\footnote{%
http://archive.ics.uci.edu/ml and https://www.dcc.fc.up.pt/$\sim$ltorgo/}. With the training and testing samples drawn as in Table 1, we used 10-fold cross-validation
to select all the parameters involved in each algorithm.
Then we implemented each algorithm independently 50 times and
calculated the  rooted mean square error (TestRMSE) of the estimator. It was also recorded the
corresponding average training time (TrainMT) for each algorithm.
For comparison of testing complexity, we recorded
the average number of hidden neurons (MSparsity) involved in LtDaHP, LtRaHP, and SVR. The simulation results are listed in Table II.

We can see from Table II that LtDaHP works well for most of the
data sets, exhibiting an almost similar or comparable generalization
performance to GPR. Both LtRaHP and SVR failed in certain data sets.

As far as the training time and testing complexity are
concerned, LtDaHP and LtRaHP significantly outperform SVR, and are better than GPR when sample size is higher than 10000. Furthermore, we can
observe that LtDaHP and LtRaHP always keep a similar training time and testing complexity.

\subsection{Real World Massive Data Experiments}

In this section we  assess the performance of LtDaHP and LtRaHP
through applying the algorithms to a real world massive data.

The problem we have applied is the household electric power consumption data set. The task is
to predict the global active power from $8$
primary features. The dataset contains $2075259$ samples, and so a real large
scale problem. We applied LtDaHP and LtRaHP to this problem by dividing the sample dataset into a training set
containing $90\%$ samples and a test set containing $10\%$
samples. 10 random partitions of the data were implemented and the results were recorded in Table III. Under such an experimental setting, the RMSE predicted by GPR is $0.053\pm 0.000$, while the deep kernel machine(DKL) can achieve $0.048\pm 0.000$, as reported in \cite{Wilson2016}.

\begin{figure}
\begin{center}
\includegraphics[scale=0.6]{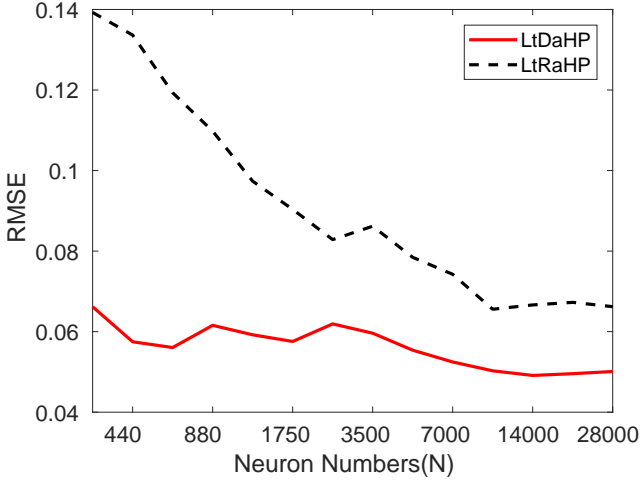}
\caption{The comparisons of test error of household electricity data
between LtDaHP and LtRaHP.}\label{fig.real}
\end{center}
\end{figure}

We only compare the performance of LtDaHP and LtRaHP because of the
extremely high computational burden of SVR for such a large scale
problem. Both algorithms were applied with neuron number $N$ varying
from $400$ to $30000$. We plot the obtained RMSEs of test error as a
function of $N$ to demonstrate the performances of LtDaHP and LtRaHP
in Fig. \ref{fig.real}. Fig. \ref{fig.real} shows that   for most
choices of $N$, LtDaHP performs much better than LtRaHP.

 To compare the performance of LtDaHP and LtRaHP further, we implemented
 the algorithms in which the
 parameters were chosen by using the 5-fold cross-validation
 method. The resultant RMSEs, TrainMT, and Msparsity are shown in
 Table III.



\begin{table}[h]
\begin{center}
\begin{tabular}{l|l|l|l}
Methods & TestRMSE & TrainMT & Msparsity  \\ \hline
LtDaHP & $0.049\pm 0.000$ & 57330s & 14032  \\
LtRaHP & $0.065\pm 0.001$ & 57610s & 14032   \\
\end{tabular}
\caption{Comparison of LtDaHP and LtRaHP on the house electric dataset}
\end{center}
\end{table}

From Table III we see that LtDaHP performs much better than LtRaHP
with respect to the TestRMSE, and they achieve a similar training
time and model sparsity.  In addition, LtDaHP achieves comparable
result to DKL, and slightly better than GPR. But it should be noted
that both DKL and GPR are lack of theoretical guarantees. However,
both LtDaHP and LtRaHP cost more computation time than DKL(which is
3600s as reported in \cite{Wilson2016}) due to an additional
procedure of parameter tuning. We believe that with some other
additional precision-promoting skills used, like ``divide and
conquer'' in \cite{Lin2016,Chang2017}, the performance of LtDaHP can
be further improved.

All these simulations and experiments support that, as a new TSL
scheme, LtDaHP outperforms LtRaHP in generalization capability and
SVR in computational complexity.  It is also comparable with GPR in
numerical ability but possesses almost optimal theoretical
guarantees.

\section{Conclusions}

In this paper, we proposed a new TSL scheme:  \textit{learning
through deterministic assignment of hidden parameters} (LtDaHP). The
 main contributions can be concluded  as follows:

\begin{itemize}

\item Borrowing an approximate solution to the classical Tamme's
problem, we  suggested to set  inner weights of FNNs as
 minimal Riesz $\tau $-energy points on the sphere and  thresholds as
equally spaced points in an interval;

\item We proved that with the suggested deterministic assignment mechanism
of the \textit{hidden parameters},  LtDaHP achieves an almost optimal
generalization bound (learning rate). In particular, it
does not   degrade the generalization capability of the
classical one-stage learning schemes very much.

\item A series of simulations and application examples were provided to
support the correctness of the theoretical assertions and the
effectiveness of the LtDaHP scheme.
\end{itemize}


Additionally, we found that the outperformance of LtDaHP over LtRaHP demonstrated in the   house electric prediction problem in Section V.C shows that LtDaHP may be
more effectively applied to practical problems, especially for large
scaled problems.
We finish this section with two additional remarks.

\begin{remark}
It should be remarked that LtDaHP involves an adjustable parameter
$K$ that may have a crucial impact for its performance. We have
suggested a criterion of specification of $K$ for the logistic
function, and the simulations in Section 4 have substantiated the
validity and effectiveness of such a criterion. This criterion is,
however, by no means universal. In other words, it is perhaps
inadequate for other activation functions. Thus, how to generally
set an appropriate K in implementation of LtDaHP is still open. We
leave it for our future study.
\end{remark}

\begin{remark}
The minimal R\textit{iesz }energy points were approximated by the
recursive zonal sphere partitioning. However, the EQSP algorithm is not robust in high-dimensional cases, which makes the corresponding LtDaHP scheme instably training high dimensional data.  We will work on data-driven methods by taking the sample distribution to determine the hidden parameters in a future work.
\end{remark}

\section*{Appendix:Proof of Theorem 1}

We divide the proof  of Theorem 1 into five parts. The first part
concerns the orthonormal system in $L^2(\mathbb B^d)$. The second
part  focuses on the ridge representations for polynomials. The
third one aims at constructing an FNN in $\mathcal
H_{\ell,n,\phi_K}$, while the fourth part pursues its approximation
ability. In the last part, we analyze the learning rate of
(\ref{Estimator}).

 \subsection{Orthonormal basis for multivariate polynomials on the
 unit ball}

 Let $G_s^\nu(t)$ be the Gegenbauer polynomial
\cite{Wang2000} with index $\nu$. It is known that the family of
polynomials $\{G_s^\nu\}_{s=0}^\infty$ is a complete orthogonal
system in the weighted space $L^2(\mathbb I,w_\nu)$ with $\mathbb
I:=[-1,1]$ and $w_\nu(t):=(1-t^2)^{\nu-\frac12}$, that is,
$$
             \int_{\mathbb I}G_{s'}^\nu(t)G_s^\nu(t)w_\nu(t)dt=\left\{\begin{array}{cc}
             0,& s'\neq s\\
             h_{s,\nu}, & s'=s
             \end{array}
             \right.,
$$
 where
 $
  h_{s,\nu}=\frac{\pi^{1/2}(2\nu)_s\Gamma(\nu+\frac12)}{(s+\nu)s!\Gamma(\nu)},
 $ and
$$
              (a)_0:=0,
              (a)_s:=a(a+1)\dots(a+s-1)=\frac{\Gamma(a+s)}{\Gamma(a)}.
$$
 Define
\begin{equation}\label{un}
              U_s:=(h_{s,d/2})^{-1/2}G_s^{d/2}, \ s=0,1,\dots.
\end{equation}
 Then,   $\{U_s\}_{s=0}^\infty$ is a
complete orthonormal system for the weighted space $L^2(\mathbb
I,w)$ with $w(t):=(1-t^2)^\frac{d-1}2$. With this, we introduce the
univariate Sobolev spaces
$$
       W^\alpha(L^2(\mathbb
       I,w)):=\left\{g:\|g\|^2_{W^\alpha(L^2(\mathbb
       I,w))}=\sum_{k=0}^\infty\left[(k+1)^\alpha
       \hat{g}_{w,k}\right]^2<\infty\right\},
$$
where $
          \hat{g}_{w,k}:=\int_{\mathbb
       I} g(t)U_k(t)w(t)dt.
$
 It is easy to see that
\begin{equation}\label{Bernstein inequality}
      \|p\|_{W^\alpha(L^2(\mathbb
       I,w))}\leq (s+1)^\alpha\|p\|_{L^2(\mathbb
       I,w)}, \qquad \forall\ p\in\mathcal P_s(\mathbb I),
\end{equation}
where $\mathcal P_s(\mathbb I)$ denotes the algebraic polynomials
defined on $\mathbb I$ of degrees at most $s$.

 Denote by $\mathbb{H}^{d-1}_j$ and $\Pi_s^{d-1}$ the class of all
spherical harmonics of degree
  $j$ and
 the class of all spherical polynomials with total degrees $j\leq
 s$, respectively. It can be found in \cite{Wang2000} that
$\Pi_s^{d-1}=\bigoplus_{j=0}^s\mathbb{H}^{d-1}_j$. Since
 the dimension of
$\mathbb{H}^{d-1}_j$ is given by
$$
                 D_j^{d-1}:=\mbox{dim } \mathbb{H}^{d-1}_j
                 =\left\{\begin{array}{ll}
                 \frac{2j+d-2}{j+d-2}{{j+d-2}\choose{j}}, & j\geq 1;\\
                 1, & j=0,
                \end{array}
                  \right.
$$
 the dimension of  $\Pi_s^{d-1}$ is $
\sum_{j=0}^sD^{d-1}_j=D_s^d\sim s^{d-1}.$ Let
$\{Y_{j,i}:i=1,\dots,D_j^{d-1}\}$ be an arbitrary orthonormal system
of $\mathbb H_j^{d-1}$.  The well known addition formula is given by
\cite{Wang2000}
\begin{equation}\label{addition}
                 \sum_{i=1}^{D_j^{d-1}}Y_{j,i}(\xi)Y_{j,i}(\eta)
                 =\frac{2j+d-2}{(d-2)\Omega_{d-1}}G_j^{\frac{d-2}2}(\xi\cdot\eta)=K_j^*(\xi\cdot\eta),
\end{equation}
where
$K_n^*(t):=\frac{2j+d-2}{(d-2)\Omega_{d-1}}G_j^{\frac{d-2}2}(t)$, $
                        \Omega_{d-1}:=\int_{\mathbb{S}^{d-1}}d\omega_{d-1}=\frac{2\pi^{\frac{d}2}}{\Gamma(\frac{d}2)},
 $
and $d\omega_{d-1}$ denotes the {aero element} of $\mathbb S^{d-1}$

For $x\in\mathbb B^d$, define
\begin{equation}\label{basis}
               P_{k,j,i}(x)=v_k\int_{\mathbb
               S^{d-1}}Y_{j,i}(\xi)U_k(x\cdot\xi)d\omega_{d-1}(\xi),
\end{equation}
where
$v_k:=\left(\frac{(k+1)_{d-1}}{2(2\pi)^{d-1}}\right)^\frac12$. Then
it follows from  \cite{Maiorov2013} that
$$
            \{P_{k,j,i}:k=0,1,\dots,s,
                j=k,k-2,\dots,\varepsilon_k,i=1,2,\dots,D_j^{d-1}\}
$$
 is an orthonormal basis for  $\mathcal P_s(\mathbb B^d)$ with $
            \varepsilon_k:=\left\{\begin{array}{cc} 0,& k\ \mbox{even},\\
                  1,&k\ \mbox{odd}\end{array}\right..
$ Based on the orthonormal system, we define the Sobolev space on
$\mathbb B^d$, denoted  by $H^r(L^2(\mathbb
       B^d))$, as the space
$$
      \left\{f:\|f\|^2_{H^r(L^2(\mathbb
       B))}=\sum_{k=0}^\infty\sum_{j\in\Xi_k}\sum_{i=1}^{D_j^{d-1}}
       \left[(k+1)^r\hat{f}_{k,j,i}\right]^2<\infty\right\},
$$
where $\hat{f}_{k,j,i}:=\int_{\mathbb
       I}f(x)P_{k,j,i}(x)dx$ and
       $\Xi_k:=\{k,k-2,\dots,\varepsilon_k\}$.

\subsection{Ridge function representation for multivariate
polynomials on the ball}

For $s\geq0$ and $N\geq 1$, let a discretization quadrature rule
\begin{equation}\label{quadrature rule}
      \mathcal Q_N(s,n):=\{(\lambda_\ell,\xi_\ell):\ell=1,\dots,n\},\quad
      \lambda_\ell>0
\end{equation}
holds exact for $\Pi_s^{d-1}$.
By the sequence of works in
\cite{Kuijlaars1998,Kuijlaars2007}, we find that all
minimal $\tau$-energy configurations with $\tau\geq d$ is a discretization quadrature rule
$\mathcal Q_N(2s,n)$.
 The following positive cubature
formula can be found in \cite{Brown2005}.

\begin{lemma}\label{CUBATURE2}
If $\tau\geq d$, then there exists a set of numbers
$\{\lambda_z\}_{z\in\mathcal {E}_\tau(\mathbb S^{d-1},n)}$ such that
$$
                 \int_{\mathbb{S}^{d-1}}P(y)d\omega(y)=\Omega_{d-1}\sum_{z\in\mathcal{E}_\tau(\mathbb S^{d-1},n)}\lambda_zP(z)
                 \ \ \mbox{for} \ \mbox{any \ \ } P\in\Pi_{2s}^{d-1}.
$$
\end{lemma}

 We then present the main tool for our analysis in the
following proposition.

\begin{proposition}\label{Proposition:ridge presentation}
Let $s\in\mathbb N$.  If $\mathcal
Q_N(2s,n)=\{\lambda_\ell,\xi_\ell\}_{\ell=1}^n$ is a discretization
quadrature rule for spherical polynomials of degree up to $2s$,
 then for arbitrary
$P\in\mathcal P_s(\mathbb B^d)$, there holds
\begin{equation}\label{ridge representation}
              P(x)=\sum_{\ell=1}^n\lambda_\ell p_\ell(\xi_\ell\cdot x),
\end{equation}
and
\begin{equation}\label{bound polynomial}
        \sum_{\ell=1}^n\lambda_\ell\|p_\ell\|^2_{W^1(L^2(\mathbb I,w))}\leq c_3\|P\|^2_{H^{\frac{d+1}2}(L^2(\mathbb
        B^d))},
\end{equation}
where
\begin{equation}\label{def.bi}
            p_\ell(t):=\sum_{k=0}^sv_k^2\int_{\mathbb
           B^d}U_k(\xi_\ell\cdot y)P(y)dyU_k(t)
\end{equation}
and $c_3$ is a constant depending only on $d$.
\end{proposition}

We postpone the proof of Proposition \ref{Proposition:ridge
presentation} to the end of this subsection and introduce the
following lemma concerning  important properties of $U_s$ at first.

\begin{lemma}\label{Lemma:tool}
Let $U_s$ be defined as  above.   Then for each $ \xi,\eta\in\mathbb
S^{d-1}$ we have
\begin{equation}\label{1}
               \int_{\mathbb B^d}U_s(\xi\cdot x)P(x)dx=0\
               \mbox{for}\
                P\in\mathcal P_{s-1}(\mathbb B^d),
\end{equation}
\begin{equation}\label{2}
              \int_{\mathbb B^d}U_s(\xi\cdot x)U_s(\eta\cdot
              x)dx=\frac{U_s(\xi\cdot\eta)}{U_s(1)},
\end{equation}
\begin{equation}\label{3}
                 K_s^*+K_{s-2}^*
                 +\dots+K_{\varepsilon_s}^*=\frac{v_s^2}{U_s(1)}U_s,
\end{equation}
\begin{equation}\label{4}
                \int_{\mathbb S^{d-1}}U_s(\xi\cdot
                x)U_s(\xi\cdot\eta)d\omega_{d-1}(\xi)=\frac{U_s(1)}{v_s^2}U_s(\eta\cdot
                x),
\end{equation}
and
\begin{equation}\label{5}
             \int_{\mathbb
                  S^{d-1}}U_k(\xi\cdot x)U_k(\xi\cdot
                  y)d\omega_{d-1}(\xi) =
                  \frac1{v_k^2}\sum_{j\in\Xi_k}\sum_{i=1}^{D_j^{d-1}}P_{k,j,i}(x)P_{k,j,i}(y).
\end{equation}
\end{lemma}

\begin{IEEEproof}
 (\ref{1})-(\ref{4}) can be found in eqs. (3.4), (3.10),
(3.16), (3.11) of \cite{Petrushev1999}, respectively. It suffices to
prove (\ref{5}). We get from (\ref{basis}) that
\begin{eqnarray*}
                 &&\sum_{j\in\Xi_k}\sum_{i=1}^{D_j^{d-1}}P_{k,j,i}(x)P_{k,j,i}(y)\\
                 &=&
                 \sum_{j\in\Xi_k}\sum_{i=1}^{D_j^{d-1}}v_k^2\int_{\mathbb
                 S^{d-1}}Y_{j,i}(\xi)U_k(x\cdot\xi)d\omega_{d-1}(\xi)\\
                 &&\int_{\mathbb
                 S^{d-1}}Y_{j,i}(\eta)U_k(y\cdot\eta)d\omega_{d-1}(\eta)\\
                 &=&
                 v_k^2\sum_{j\in\Xi_k}\int_{\mathbb
                 S^{d-1}}U_k(x\cdot\xi)\int_{\mathbb
                 S^{d-1}}U_k(y\cdot\eta)\\
                 &\times&\sum_{i=1}^{D_j^{d-1}}Y_{j,i}(\xi)Y_{j,i}(\eta)d\omega_{d-1}(\xi)d\omega_{d-1}(\eta).
\end{eqnarray*}
 Thus,   the addition formula
(\ref{addition}) yields
\begin{eqnarray*}
                 &&\sum_{j\in\Xi_k}\sum_{i=1}^{D_j^{d-1}}P_{k,j,i}(x)P_{k,j,i}(y)
                 =v_k^2\sum_{j\in\Xi_k}
                 \int_{\mathbb
                 S^{d-1}}\\
                 &\times&
                 U_k(x\cdot\xi)\int_{\mathbb
                 S^{d-1}}U_k(y\cdot\eta)K^*_j(\xi\cdot\eta)d\omega_{d-1}(\xi)d\omega_{d-1}(\eta).
\end{eqnarray*}
The above equality together with (\ref{3}) and (\ref{4}) implies
\begin{eqnarray*}
                 &&
                 \sum_{j\in\Xi_k}\sum_{i=1}^{D_j^{d-1}}P_{k,j,i}(x)P_{k,j,i}(y)
                  =
                v_k^2\int_{\mathbb
                 S^{d-1}}U_k(x\cdot\xi)\\
                 &\times&\int_{\mathbb
                 S^{d-1}}U_k(y\cdot\eta)\sum_{j\in\Xi_k}K_j^*(\xi\cdot\eta)d\omega_{d-1}(\xi)d\omega_{d-1}(\eta)\\
                 &=&
                 \frac{v_k^4}{U_k(1)}\int_{\mathbb
                 S^{d-1}}U_k(x\cdot\xi)\\
                 &\times&\int_{\mathbb
                 S^{d-1}}U_k(y\cdot\eta)U_k(\xi\cdot\eta)d\omega_{d-1}(\xi)d\omega_{d-1}(\eta)\\
                 &=&
                 v_k^2\int_{\mathbb
                  S^{d-1}}U_k(\xi\cdot x)U_k(\xi\cdot
                  y)d\omega_{d-1}(\xi).
\end{eqnarray*}
This competes the proof of Lemma \ref{Lemma:tool}.
\end{IEEEproof}

Based on (\ref{5}) and the well known Aronszajn Theorem, it is easy to construct a reproducing kernel
of $\mathcal P_s(\mathbb B^d)$.

\begin{lemma}\label{Lemma: reproducing kerne}
The space $(\mathcal P_s(\mathbb B^d), \langle
\cdot,\cdot\rangle_{L^2(\mathbb B^d)})$ is a reproducing kernel
Hilbert space with the   reproducing kernel
\begin{equation}\label{reproducing kernel}
                  K_s(x,y):=\sum_{k=0}^sv_k^2\int_{\mathbb
                  S^{d-1}}U_k(\xi\cdot x)U_k(\xi\cdot
                  y)d\omega_{d-1}(\xi).
\end{equation}
\end{lemma}

By the help of these lemmas, we are in a position to prove
Proposition \ref{Proposition:ridge presentation}.

\begin{IEEEproof}[Proof of Proposition \ref{Proposition:ridge presentation}]
Due
to Lemma \ref{Lemma: reproducing kerne}, for arbitrary $P\in
\mathcal P_s(\mathbb B^d)$,
 there
holds
\begin{eqnarray*}
           &&P_s(x)=\int_{\mathbb B^d}K_s(x,y)P_s(y)dy\\
           &=&
           \int_{\mathbb
           B^d}\sum_{k=0}^sv_k^2\int_{\mathbb S^{d-1}}U_k(\xi\cdot
           x)U_k(\xi\cdot y)d\omega_{d-1}(\xi)P_s(y)dy.
\end{eqnarray*}
Since  $\mathcal Q_N(2s,N)=\{\lambda_\ell,\xi_\ell\}_{\ell=1}^N$ is
a discretization quadrature rule for spherical polynomials of degree
up to $2s$, we have
$$
          P_s(x)=\sum_{k=0}^sv_k^2\sum_{\ell=1}^N\lambda_\ell\int_{\mathbb
           B^d}U_k(\xi_\ell\cdot y)P_s(y)dyU_k(\xi_\ell\cdot x).
$$
 This proves (\ref{ridge representation}).
To derive (\ref{bound polynomial}), we use (\ref{def.bi}), (\ref{5})
and  $v_k\leq c_3k^{\frac{d-1}2}$ for some $c_3\geq 1$
 and obtain
\begin{eqnarray*}
  && \sum_{\ell=1}^N\lambda_\ell\|p_\ell\|^2_{W^1(L^2(\mathbb I,w))}\\
  &=&
  \sum_{k=0}^\infty (k+1)^2\sum_{\ell=1}^N\lambda_\ell\left(\int_{\mathbb
      I} p_\ell(t)U_k(t) w(t)dt\right)^2\\
      &=&
      \sum_{k=0}^\infty (k+1)^2\sum_{\ell=1}^N\lambda_\ell\\
      &\times&
      \left(\int_{\mathbb
      I}\sum_{k'=0}^sv_{k'}^2\int_{\mathbb
           B^d}U_{k'}(\xi_\ell\cdot y)P(y)dyU_{k'}(t)U_k(t) w(t)dt\right)^2\\
           &=&
     \sum_{k=0}^s (k+1)^2 \sum_{\ell=1}^N\lambda_\ell\left( v_k^2\int_{\mathbb
           B^d}U_k(\xi_\ell\cdot y)P(y)dy \right)^2\\
           &=&
       \sum_{k=0}^s(k+1)^2 v_k^4\sum_{\ell=1}^N\lambda_\ell \int_{\mathbb
           B^d}P(y) \\
           &&
           \int_{\mathbb
           B^d}P(z)U_k(\xi_\ell\cdot y)U_k(\xi_\ell\cdot z) dydz\\
        &=&
        \sum_{k=0}^s(k+1)^2 v_k^4\int_{\mathbb
           B^d}P(y)\int_{\mathbb
           B^d}P(z) \\
           &&
           \int_{\mathbb S^{d-1}}U_k(\xi\cdot y)U_k(\xi\cdot z)d\omega_{d-1}(\xi)
           dydz\\
           &=&
           \sum_{k=0}^s(k+1)^2 v_k^2\sum_{j\in\Xi_k}\sum_{i=1}^{D_k^{d-1}}\int_{\mathbb
           B^d}P(y)P_{k,j,i}(y)dy\int_{\mathbb
           B^d}P(z)P_{k,j,i}(z)dz\\
           &\leq&
           c_3\sum_{k=0}^s(k+1)^{d+1}\sum_{j\in\Xi_k}\sum_{i=1}^{D_k^{d-1}}\hat{P}_{k,j,i}
            =
           c_3\|P_s\|_{H^{\frac{d+1}2}(L^2(\mathbb B^d))}^2.
\end{eqnarray*}
This completes the proof of Proposition \ref{Proposition:ridge
presentation}.
\end{IEEEproof}

\subsection{Constructing neural networks}

Before constructing the neural networks, we at first present a univariate Jackson-type  error estimate for
FNNs with sigmoidal activation function, which can be found in \cite{Chui1994,Cao2009}.
For $\ell\in\mathbb N$ and $\ell\geq 2$, let $t_j=-\frac12+\frac{j}\ell$ with $0\leq j\leq \ell$ be the equally spaced
points on $\mathbb J:=[-1/2,1/2]$.
Define
$$
       \Phi_\ell^1:=\sum_{j=0}^{\ell-1}c_i\phi(K(t-t_j)).
$$
Then, it can be found in   \cite[Theorem 1]{Cao2009} the following error estimate.
 \begin{lemma}\label{Lemma:distance}
Let $m\in\mathbb N$, $\ell\in\mathbb N$ with $\ell\geq 2$ and $\phi(\cdot)$, $K$ satisfy (\ref{choose K}) and (\ref{definition L}), respectively.   Then there exists a $g^*\in  \Phi_\ell^1$ such that
$$
       \|p-g^*\|_{L^2(\mathbb J)} \leq \tilde{c} (\ell^{-1}+\ell m^{-2/(2r+d)})\|p\|_{W^1(L_2(\mathbb J,w))}.
$$
where $\tilde{c}$ is an  absolute constant.
\end{lemma}

Now, we proceed our construction.
  For arbitrary $f\in
W^r(L^2(\mathbb
       B^d_{1/2}))$,
we at first extend $f$ to a function $f^*$ defined on $\mathbb B^d$
  \cite{Petrushev1999}  such
that $f^*$ vanishes outside of $\mathbb B_{3/4}^d$ and
\begin{equation}\label{extension}
       \|f^*\|_{H^{r}(L^2(\mathbb B^d))}\leq  c'
       \|f\|_{W^{r}(L^2(\mathbb B^d_{1/2}))},
\end{equation}
where $c'>0$ is a constant depending only on $d$. If  we define
\begin{equation}\label{def.p}
      P_s(x)=\sum_{k=0}^s\sum_{j\in\Xi_k}\sum_{i=1}^{D_j^{d-1}}\hat{f^*}_{k,j,i}P_{k,j,i}(x),
\end{equation}
Then
\begin{eqnarray}\label{Jackson for P}
     &&\|f^*-P_s\|^2_{L^2(\mathbb B^d)}
      =
     \sum_{k=s+1}^\infty\sum_{j\in\Xi_k}\sum_{i=1}^{D_j^{d-1}}|\hat{f^*}_{k,j,i}|^2\\
     &\leq&
     s^{-2r}\sum_{k=s+1}^\infty\sum_{j\in\Xi_k}\sum_{i=1}^{D_j^{d-1}}
     [(k+1)^{r}|\hat{f^*}_{k,j,i}|]^2\nonumber\\
     &\leq&
     s^{-2r} \|f^*\|^2_{W^r(L^2(\mathbb
       B^d))}\leq (c')^2s^{-2r}\|f\|^2_{W^r(L^2(\mathbb
       B^d_{1/2}))}.
\end{eqnarray}
Due to Proposition \ref{Proposition:ridge presentation}, we get
\begin{equation}\label{d p}
              P_s(x)=\sum_{\ell=1}^N\lambda_\ell p_\ell(\xi_\ell\cdot x),
\end{equation}
where $p_\ell$ is defined by (\ref{def.bi}). We then aim at
constructing the neural network based on the  following lemma
\cite{Petrushev1999}.

\begin{lemma}\label{Lemma:functional lemma}
 Let $H$ be a Hilbert space with norm $\|\cdot\|$ and let
$A,B\subset H$ be finite dimensional linear subspaces of $H$ with
$\mbox{dim}A\leq\mbox{dim}B.$ If there exists a $\delta\in(0,1/2)$
  such that
$$
                 \sup_{x\in A,\ \|x\|\leq1}\inf_{y\in
                 B}\|x-y\|\leq\delta,
$$
then there is a constant $c$ depending only on $\delta$ and a linear
operator $L:A\rightarrow B$ such that for every $x\in A,$
$$
                    \|Lx-x\|\leq c\inf_{y\in B}\|x-y\|,
$$
and
$$
                     Lx-x\bot A\ (Lx-x\ \mbox{is orthogonal to}\ A).
$$
\end{lemma}

To use Lemma \ref{Lemma:functional lemma}, we define
$$
\psi_{k,j}(t):=\left\{\begin{array}{cc}
      T_{\tau,t_k,t_{k+1}}(t),& t\in \mathbb J,\\
       t^j,& t\in\mathbb I\backslash \mathbb
       J\end{array}\right.
$$
 Then we get an $n^2$-dimensional  linear space
\begin{equation}\label{Set initial}
       \Phi_{n}:=\left\{ \sum_{k=0}^{n-1}\sum_{j=0}^{n-1}c_{k,j}\psi_{k,j}(t):c_{k,j}\in\mathbb
       R\right\}.
\end{equation}
Take for $H$ the Hilbert space $L^2(\mathbb I,w)$,  $A=\mathcal
P_s(\mathbb I)$ and $B$ the space $\Phi_{\ell}^1$ defined above. Set
$\ell=4\sqrt{\tilde{c}+1}s$ in Lemma \ref{Lemma:functional lemma}. It
then follows from Lemma \ref{Lemma:distance} that
\begin{eqnarray}\label{jackson for media}
      &&\sup_{p\in A}\inf_{g\in B} \|p-g\|_{L_2(\mathbb I,w)}
      \leq \sup_{p\in A} \|p-g^*\|_{L_2(\mathbb J,w)}\\
      &\leq&
      \sup_{p\in A} \|p-g^*\|_{L_2(\mathbb J)}
       \leq
      \tilde{c} (\ell^{-1}+\ell m^{-2/(2r+d)})\|p\|_{W^1(L_2(\mathbb J,w))}.\nonumber
\end{eqnarray}
But the  Bernstein inequality (\ref{Bernstein inequality}) shows
$$
    \sup_{p\in A}\inf_{g\in B} \|p-g\|_{L_2(\mathbb I,w)}
    \leq
    \tilde{c} (\ell^{-1}+\ell m^{-2/(2r+d)})(s+1)\|p\|_{L_2(\mathbb I,w)}.
$$
Since $\ell=\bar{c}m^{-1/(2r+d)}$, we have
$$
          \sup_{p\in A}\inf_{g\in B} \|p-g\|_{L_2(\mathbb I,w)}
          \leq
          \frac14.
$$
Thus, the condition of Lemma \ref{Lemma:functional lemma} is
satisfied. Then, it follows from Lemma \ref{Lemma:functional lemma}
and (\ref{jackson for media})
 that for arbitrary $\ell=1,2,\dots N$,  there is a
\begin{equation}\label{initial q}
         q_l(t)=\left\{\begin{array}{cc}\sum_{k=0}^{n-1}a_{k,\ell}
         \phi(K(t-t_j)),& t\in \mathbb J,\\
       \sum_{j=0}^{n-1} a'_{j,\ell} t^j,& t\in\mathbb I\backslash \mathbb
       J\end{array}\right.\in\Phi_n
\end{equation}
for some sequences $\{a'_{j,\ell}\}_{j=0}^{n-1}$,
$\{a_{k,\ell}\}_{k=1}^{n-1}$ such that
$$
     \|q_\ell-p_\ell\|^2_{L_2(\mathbb I,w)}
     \leq
     c\tilde{c}\tilde{c} (\ell^{-1}+\ell m^{-2/(2r+d)})\|p\|_{W^1(L_2(\mathbb J,w))}.
$$
and
\begin{equation}\label{resuidal}
          q_\ell(t)-p_\ell(t)=\sum_{k=s+1}^\infty\hat{q}_\ell(k,w)U_k(t).
\end{equation}
The above two estimates  show
\begin{eqnarray}\label{importantestimate}
              &&\sum_{k=s+1}^\infty|\hat{q_\ell}(k,w)|^2
              \leq
               c_6n^{-2} \|p_\ell\|^2_{W^1(L_2(\mathbb I,w))},
\end{eqnarray}
 where $c_6:= c\tilde{c}+\frac{c\tilde{c}}{16(\tilde{c}+1)}.$
With these helps, we define
\begin{eqnarray}\label{def.Q}
         &&Q(x):=\sum_{\ell=1}^N\lambda_\ell q_\ell(\xi_\ell\cdot x)\nonumber\\
         &=&\sum_{\ell=1}^N\lambda_\ell
         \left\{\begin{array}{cc}\sum_{k=0}^{n-1}a_{k,\ell}
         T_{\tau,t_k,t_{k+1}}(\xi_\ell\cdot x),& \xi_\ell\cdot x\in \mathbb J,\\
       \sum_{j=0}^{n-1} a'_{j,\ell} (\xi_\ell\cdot x)^j,& \xi_\ell\cdot x \in\mathbb I\backslash \mathbb
       J.\end{array}\right.
\end{eqnarray}
Since $\xi_\ell\in\mathbb S^{d-1}$ for $\ell=1,\dots,N$,
$x\in\mathbb B^d_{1/2}$ implies $\xi_\ell\cdot x\in \mathbb J$ and
thus
\begin{equation}\label{def.Q1}
      Q^*(x)=\sum_{\ell=1}^N\lambda_\ell\sum_{k=0}^{n-1}a_{k,\ell}
         T_{\tau,t_k,t_{k+1}}(\xi_\ell\cdot x),\qquad \forall\
         x\in\mathbb B^d_{1/2}.
\end{equation}

\subsection{Approximation error analysis}

Based on the previous construction, we obtain the following approximation error estimate.
\begin{lemma}\label{Lemma:APPROXIMATION ERROR}
Let $0\leq r\leq\frac{d+1}2.$ If $f_\rho\in W^r_2$ and $n\sim\ell^{d-1}$, then there is a function $\varphi\in\mathcal
H_{l,n,\phi_K}$ such that
$$
           \|f_\rho-\varphi\|_{L^2(\mathbf B^d)}\leq C(nl)^{- r/d},
$$
where $C$ is a constant depending only on $r$, $d$, $\phi$ and
$f_\rho$.
\end{lemma}

We postpone the proof of the above lemma to the last of this subsection.
We have from (\ref{d p}) and (\ref{resuidal}) that
\begin{eqnarray*}
           &&Q(x)-P_s(x)
           =
           \sum_{\ell=1}^N\lambda_\ell\left(q_\ell(\xi_\ell\cdot x)-p_\ell(\xi_\ell\cdot
           x)\right)\\
           &=&
           \sum_{\ell=1}^N\lambda_\ell\sum_{k=s+1}^\infty\hat{q_\ell}(k,w)U_k(\xi_\ell\cdot x).
\end{eqnarray*}
Denote
$$
           Q_k(x):=\sum_{\ell=1}^N\lambda_\ell\hat{q_\ell}(k,w)U_k(\xi_\ell\cdot
           x).
$$
The following lemma derives the bound of $\|Q_k\|_{L^2(\mathbb
B^d)}$.

\begin{lemma}\label{Lemma:bound Qk}
Let $Q_k$ be defined above. Then,
\begin{equation}\label{Q ball}
    \|Q_k\|^2_{L^2(\mathbb B^{d})}
    \leq
     v_k^{-2}\sum_{\ell=1}^N\lambda_\ell|\hat{q_\ell}(k,w)|^2.
\end{equation}
\end{lemma}

\begin{IEEEproof}
 From Lemma \ref{Lemma:tool}, we obtain
\begin{eqnarray*}
          &&\|Q_k\|_{L^2(\mathbb B^d)}^2
           =
           \sum_{k'=0}^\infty\sum_{j\in\Xi_k}\sum_{i=1}^{D_j^{d-1}}\left(\int_{\mathbb
          B^d}Q_k(y)P_{i,j,k'}(y)dy\right)^2\\
          & =&
          \sum_{k'=0}^\infty\sum_{j\in\Xi_k}\sum_{i=1}^{D_j^{d-1}}\\
          &&\left(\int_{\mathbb
          B^d}Q_k(y)v_{k'}\int_{\mathbb S^{d-1}}Y_{j,i}(\xi)U_{k'}(y\cdot
          \xi)d\omega_{d-1}(\xi)dy\right)^2\\
          &=&
          \sum_{k'=0}^\infty v_{k'}^2\sum_{j\in\Xi_k}\sum_{i=1}^{D_j^{d-1}}
          \left(\int_{\mathbb
          S^{d-1}}Y_{j,i}(\xi)\right.\\
          &&
          \left.\int_{\mathbb
          B^d}\sum_{\ell=1}^N\lambda_\ell\hat{q_\ell}(k,w)U_k(\xi_\ell\cdot y)
          U_{k'}(y\cdot
          \xi)dyd\omega_{d-1}(\xi)\right)^2\\
          &=&
          v_k^2\sum_{j\in\Xi_k}\sum_{i=1}^{D_j^{d-1}}
          \left(\int_{\mathbb
          S^{d-1}}Y_{j,i}(\xi)\sum_{\ell=1}^N\lambda_\ell\hat{q_\ell}(k,w)\right.\\
          &&\left.\int_{\mathbb
          B^d}U_k(\xi_\ell\cdot y)U_k(y\cdot
          \xi)dyd\omega_{d-1}(\xi)\right)^2\\
          &=&
          \frac{v_k^2}{(U_k(1))^2}\sum_{j\in\Xi_k}\sum_{i=1}^{D_j^{d-1}}\\
          &&
          \left(\sum_{\ell=1}^N\lambda_\ell\hat{q_\ell}(k,w)
          \int_{\mathbb S^{d-1}}Y_{j,i}(\xi)U_k(\xi_\ell\cdot
          \xi)d\omega_{d-1}(\xi)\right)^2.
\end{eqnarray*}
Since (\ref{addition}) yields
$$
        \int_{\mathbb S^{d-1}}
        Y_{j,i}(\eta)K_{j'}^*(\xi\cdot\eta)d\omega_{d-1}(\eta)=
        \left\{\begin{array}{cc} Y_{j,i}(\xi),& \mbox{if}\  j'=j,\\
        0,&\mbox{if}\ j'\neq j \end{array}\right.,
$$
which together with (\ref{3}) implies
\begin{eqnarray*}
         &&\frac{v_k^2}{U_k(1)}\int_{\mathbb S^{d-1}}Y_{j,i}(\xi)U_k(\xi_\ell\cdot
          \xi)d\omega_{d-1}(\xi)\\
          &=&
          \int_{\mathbb S^{d-1}}\sum_{j'\in \Xi_k} K_{j'}^*(\xi_\ell\cdot
          \xi)Y_{j,i}(\xi)d\omega_{d-1}(\xi)\\
          &=&
          \sum_{j'\in \Xi_k} \int_{\mathbb S^{d-1}}K_{j'}^*(\xi_\ell\cdot
          \xi)Y_{j,i}(\xi)d\omega_{d-1}(\xi)=Y_{j,i}(\xi_\ell).
\end{eqnarray*}
 Thus,
\begin{eqnarray*}
          &&\|Q_k\|_{L^2(\mathbb B^d)}^2
           =
           v_k^{-2}\sum_{j\in\Xi_k}\sum_{i=1}^{D_j^{d-1}}
          \left(\sum_{\ell=1}^N\lambda_\ell\hat{q_\ell}(k,w)
           Y_{j,i}(\xi_\ell)\right)^2\\
           &=&
           v_k^{-2}
           \sum_{\ell=1}^N\lambda_\ell\hat{q_\ell}(k,w)
           \sum_{\ell'=1}^N\lambda_{\ell'}\hat{q_{\ell'}}(k,w)
           \sum_{j\in\Xi_k}\sum_{i=1}^{D_j^{d-1}}Y_{j,i}(\xi_\ell)Y_{j,i}(\xi_{\ell'}).
\end{eqnarray*}
Due to (\ref{addition}) and (\ref{3}), there holds
$$
       \sum_{j\in\Xi_k}\sum_{i=1}^{D_j^{d-1}}Y_{j,i}(\xi_\ell)Y_{j,i}(\xi_{\ell'})
       =
       \sum_{j\in\Xi_k}K_j^*(\xi_\ell\cdot \xi_{\ell'})
       =\frac{v_k^2}{U_k(1)}U_k(\xi_\ell\cdot \xi_{\ell'}).
$$
Noting further that $\mathcal
Q_N(2s,N)=\{\lambda_\ell,\xi_\ell\}_{\ell=1}^N$ is a discretization
quadrature rule for spherical polynomials of degree up to $2s$,  it
follows from  H\"{o}lder's inequality that
\begin{eqnarray}\label{media 111}
    &&\|Q_k\|_{L^2(\mathbb B^d)}^2
    =\frac{1}{U_k(1)}
    \sum_{\ell=1}^N\lambda_\ell\hat{q_\ell}(k,w)
    \sum_{\ell'=1}^N\lambda_{\ell'}\hat{q_{\ell'}}(k,w)U_k(\xi_\ell\cdot
    \xi_{\ell'}) \nonumber \\
    &=&
    \frac{1}{U_k(1)}
    \sum_{\ell=1}^N\lambda_\ell\hat{q_\ell}(k,w)Q_k(\xi_\ell)\\
    &\leq&
    \frac{1}{U_k(1)}\left(\sum_{\ell=1}^N\lambda_\ell|\hat{q_\ell}(k,w)|^2\right)^{1/2}
    \|Q_k\|_{L^2(\mathbb S^{d-1})}.
\end{eqnarray}
According to (\ref{4}), we obtain
\begin{eqnarray*}
   &&\|Q_k\|_{L^2(\mathbb S^{d-1})}^2
    =
   \int_{\mathbb S^{d-1}}
   \sum_{\ell=1}^N\lambda_\ell\hat{q_\ell}(k,w)U_k(\xi_\ell\cdot \xi)\\
   &&
   \sum_{\ell'=1}^N\lambda_{\ell'}\hat{q_{\ell'}}(k,w)U_k(\xi_{\ell'}\cdot
   \xi)d\omega_{d-1}(\xi)\\
   &=&
   \frac{U_k(1)}{v_k^2}\sum_{\ell=1}^N\lambda_\ell\hat{q_\ell}(k,w)
   \sum_{\ell'=1}^N\lambda_{\ell'}\hat{q_{\ell'}}(k,w)U_k(\xi_{\ell'}\cdot\xi_\ell)\\
   &=&
    \frac{U_k(1)}{v_k^2}\sum_{\ell=1}^N\lambda_\ell\hat{q_\ell}(k,w)Q_k(\xi_\ell)\\
    &\leq&
   \frac{U_k(1)}{v_k^2}\left(\sum_{\ell=1}^N\lambda_\ell|\hat{q_\ell}(k,w)|^2\right)^{1/2}
   \left(\sum_{\ell=1}^N\lambda_\ell (Q_k(\xi_\ell))^2\right)^{1/2}.
\end{eqnarray*}
But
$$
           \sum_{\ell=1}^N\lambda_\ell (Q_k(\xi_\ell))^2=\|Q_k\|^2_{L^2(\mathbb
           S^{d-1})}.
$$
Therefore,
\begin{equation}\label{Q sphere}
    \|Q_k\|_{L^2(\mathbb S^{d-1})}
    \leq
     \frac{U_k(1)}{v_k^2}\left(\sum_{\ell=1}^N\lambda_\ell|\hat{q_\ell}(k,w)|^2\right)^{1/2}.
\end{equation}
Inserting (\ref{Q sphere}) into (\ref{media 111}), we get
$$
    \|Q_k\|^2_{L^2(\mathbb B^{d})}
    \leq
     v_k^{-2}\sum_{\ell=1}^N\lambda_\ell|\hat{q_\ell}(k,w)|^2.
$$
This completes the proof of Lemma \ref{Lemma:bound Qk}.
\end{IEEEproof}

Now we proceed the proof of Lemma \ref{Lemma:APPROXIMATION ERROR}.
\begin{IEEEproof} It follows from
(\ref{1}) that
\begin{equation}\label{l2norm}
            \| Q-P_s\|_{L^2(\mathbb
            B^d)}^2=\sum_{k=s+1}^\infty\|Q_k\|_{L^2(\mathbb
            B^d)}^2.
\end{equation}
Plugging (\ref{Q ball}) into (\ref{l2norm}), we get
\begin{eqnarray*}
            &&\| Q-P_s\|_{L^2(\mathbb
            B^d)}^2\leq  \sum_{k=s+1}^\infty v_k^{-2}\sum_{\ell=1}^N\lambda_\ell
          |\hat{q_\ell}(k,w)|^2\\
          &\leq&
           s^{1-d}\sum_{k=s+1}^\infty \sum_{\ell=1}^N\lambda_\ell
          |\hat{q_\ell}(k,w)|^2.
\end{eqnarray*}
This together with (\ref{importantestimate}) shows
$$
            \| Q-P_s\|_{L^2(\mathbb
            B^d)}^2\leq c_6s^{1-d} n^{-2} \sum_{\ell=1}^N\lambda_\ell   \|p_\ell\|^2_{W^1(L_2(\mathbb I,w))}.
$$
Hence, we obtain from (\ref{bound polynomial}) that
$$
   \| Q-P_s\|_{L^2(\mathbb
            B^d)}^2\leq  c_3c_6s^{1-d}n^{-2} \|P_s\|^2_{W^{\frac{d+1}2}(L^2(\mathbb
        B^d))}.
$$
This together with $n=4\sqrt{\tilde{c}+1}s$ yields
$$
       \| Q-P_s\|_{L^2(\mathbb
            B^d)}\leq
             c_7n^{-\frac{d+1}2}\|P_s\|_{W^{\frac{d+1}2}(L^2(\mathbb
        B^d))}
$$
with $c_7=\sqrt{c_3c_6(4\sqrt{\tilde{c}+1})^{d-1}}$. Then,
(\ref{def.p}) and (\ref{extension}) yield
\begin{eqnarray*}
       &&\| Q-P_s\|_{L^2(\mathbb
            B^d)}\leq
             c_7n^{-\frac{d+1}2}\|f^*\|_{W^{\frac{d+1}2}(L^2(\mathbb
        B^d))}\\
        &\leq&
         c'c_7n^{-\frac{d+1}2}\|f\|_{W^{\frac{d+1}2}(L^2(\mathbb
        B^d_{1/2}))}
\end{eqnarray*}
Noting further (\ref{Jackson for P}) and $0<r\leq\frac{d+1}2$, we
have
\begin{eqnarray*}
    && \| f^*-Q\|_{L^2(\mathbb
            B^d)} \leq \|f^*-P_s\|_{L^2(\mathbb
            B^d)}+\|P_s-Q\|_{L^2(\mathbb
            B^d)}\\
            &\leq&
       c'c_7n^{-\frac{d+1}2}\|f\|_{W^{\frac{d+1}2}(L^2(\mathbb
        B^d_{1/2}))}+ c's^{-r} \|f^*\|_{W^r(L^2(\mathbb
       B^d_{1/2}))}\\
       &\leq&
        c_8n^{-r} \|f^*\|_{W^r(L^2(\mathbb
       B^d_{1/2}))}
\end{eqnarray*}
with $c_{8}=c'(c_7+c_0^{-r})$. Hence
$$
       \|f^*-Q\|_{L^2(\mathbb
            B^d_{1/2})}\leq  \| f^*-Q\|_{L^2(\mathbb
            B^d)}\leq c_8n^{-r} \|f^*\|_{W^r(L^2(\mathbb
       B^d_{1/2}))}.
$$
This completes the proof of Lemma \ref{Lemma:APPROXIMATION ERROR} by
scaling.
\end{IEEEproof}

\subsection{Proof of Theorem 1}

To prove  Theorem 1, we need   the
following Lemma \ref{ORACLE}, which was proved in \cite[Theorem
11.3]{Gyorfi2002}.

\begin{lemma}\label{ORACLE}
Let $H_m$ be a  $u$-dimensional linear space. Define the estimate
$f_{m}$ by
$$
         f_m:=\pi_M \widetilde{f}_m\quad\mbox{where}\quad
         \widetilde{f}_m=\arg\min_{f\in
         H_m}\frac1m\sum_{i=1}^m|f(x_i)-y_i|^2.
$$
Then
$$
          \mathbf E_{\rho^m}\left\{\|f_m-f_\rho\|_\rho^2\right\}\leq
          CM^2\frac{u\log m}{m}+8\inf_{f\in H_n}\|f_\rho-f\|_\rho^2,
$$
for some universal constant $c$.
\end{lemma}

Now we proceed the proof of Theorem 1. For the upper bound, we
combine Lemma \ref{Lemma:APPROXIMATION ERROR} with Lemma \ref{ORACLE}. It
can be found from the definition that $\mathcal H_{\ell,n,\phi_K}$ is
an $\ell n$-dimensional linear space.Therefore, Lemma \ref{ORACLE}
implies that
$$
          \mathbf E_{\rho^m}\left\{\|\pi_Mf_{{\bf z},\ell,n,\phi_K}-f_\rho\|_\rho^2\right\}\leq
          CM^2\frac{\ell n\log m}{m}+8\inf_{f\in \mathcal
          H_{\ell,n,\phi_K}}\|f_\rho-f\|_\rho^2.
$$
Furthermore, it follows from Lemma \ref{Lemma:APPROXIMATION ERROR} that
$$
           \inf_{f\in \mathcal
          H_{\ell,n,\phi_K}}\|f_\rho-f\|_\rho^2
          \leq D_{\rho_X}^2 \inf_{f\in \mathcal
          H_{\ell,n,\phi_K}}\|f_\rho-f\|_{L^2(\mathbf B^d)}^2\leq
          C(n\ell)^ {-r/d},
$$
provided $n\sim \ell^{d-1}$ and $f_\rho\in W^r_2$ with
$0<r\leq(d+1)/2$. Therefore, the upper bound  is deduced by setting
$n\ell=m^{d/(2r+d)}$. The lower bound   can be found
  from \cite[Theorem 3.2]{Gyorfi2002}. This completes the proof of
  Theorem 1.

\section*{Acknowledgement}
The authors
would like to thank four anonymous referees for their constructive
suggestions. The research was partly supported by   the National Natural Science
Foundations of China (Grants Nos.61876133,11771021).   Authors contributed
equally to this paper and are listed alphabetically.

\end{document}